\documentclass[10pt,twoside,leqno]{article}

\usepackage{amsfonts}
\usepackage{epsfig}
\usepackage{subfig}
\usepackage{amssymb,amsmath,bm}
\usepackage{amsthm}
\usepackage{mathrsfs}
\usepackage{graphicx}
\usepackage{booktabs} 
\usepackage{array} 
\usepackage{paralist} 
\usepackage{verbatim} 
\usepackage{algorithm2e}
\usepackage{multirow}
\usepackage{amsmath} 
\usepackage{amssymb} 
\usepackage{graphicx}
\usepackage{framed}
\usepackage{float}
\usepackage{enumerate}
\usepackage{bm}
\usepackage{subfig} 

\def\R{\in \mathbb{R}}

\def\lv{\Vert}
\def\rv{\Vert}

\def\dx{\frac{\partial}{\partial \bm{x}_i}}

\newcommand{\tr}{^{\top}}

\newtheorem{thm_pca}{Theorem}

\newtheorem{lem_pca}{Lemma}

\renewcommand{\arraystretch}{0.8}

\begin{document}

\begin{center}
	
{\Large \textbf{Subspace clustering of high-dimensional data:\\a predictive approach}}

\bigskip

BRIAN MCWILLIAMS \& GIOVANNI MONTANA\footnote{Corresponding author. Email: {\tt g.montana@ic.ac.uk}}

{\footnotesize Department of Mathematics,
Imperial College London, London, SW7 2AZ, UK.}\\
\end{center}

\begin{abstract}
In several application domains, high-dimensional observations are collected and then analysed in search for naturally occurring data clusters which might provide further insights about the nature of the problem. In this paper we describe a new approach for partitioning such high-dimensional data. Our assumption is that, within each cluster, the data can be approximated well by a linear subspace estimated by means of a principal component analysis (PCA). The proposed algorithm, Predictive Subspace Clustering (PSC) partitions the data into clusters while simultaneously estimating cluster-wise PCA parameters. The algorithm minimises an objective function that depends upon a new measure of influence for PCA models. A penalised version of the algorithm is also described for carrying our simultaneous subspace clustering and variable selection. The convergence of PSC is discussed in detail, and extensive simulation results and comparisons to competing methods are presented. The comparative performance of PSC has been assessed on six real gene expression data sets for which PSC often provides state-of-art results. 
\end{abstract}

\section{Introduction} \label{introduction}

In recent years, vast amounts of digital information has been generated which continues to grow ever more rapidly. 
For instance, within the realm of genomic research, high-throughput microarray technologies provide quantitative measurements of tens of thousands of genes for each biological entity under observation, such as a sample tissue; these measurements are then analysed to identify gene expression patterns that are predictive of a clinical outcome or to detect sub-populations. A number of other technologies, such as those developed for digital imaging, surveillance, and e-commerce, amongst many others, all generate observations which can be seen as random vectors living in a large dimensional space.  

Very often, even though the available observations are high dimensional, it has been found that their intrinsic dimensionality is in fact much smaller. For instance, although the number of pixels or voxels in a digital image can be extremely large, it is often the case that only a few important dimensions are able to capture some salient aspects of the available images, and those are sufficient to detect meaningful patterns. In genomics, despite the fact that the expression levels of tens of thousands of genes are being measured, a much smaller number of dimensions may suffice to identify different underlying biological processes which characterise the available samples.   

The abundance of such high-dimensional data and the fact that their low-dimensional representations are often interpretable and informative has caused projection-based dimensionality reduction techniques to become popular. When each data point is treated as an independent realisation from a given distribution with support in a low-dimensional subspace, Principal Component Analysis (PCA) is commonly used to recover the low-dimensional representation of the data \cite{Jolliffe}. PCA and related techniques, however, assume that the available data points are drawn from a single underlying distribution that is typical of the population from which the observations have been obtained. Whereas this can be a valid assumption in some cases, there are also many applications in which the underlying population is suspected to be highly heterogeneous; in those cases, each observation may have been drawn from one of many alternative distributions, and a single low-dimensional representation of the data would not be sufficient. 

Partitioning the sample data points according to the low-dimensional subspace that best describes them has become an important research question in various domains. In cancer genomics, for instance, the biological tissues for which high-dimensional gene expression profiles have been observed may be representative of a number of cancer sub-types; the identification of these sub-populations is important because each one of them may be associated with a distinct clinical outcome, such as life expectancy after therapy \cite{Yeoh2002}. In the analysis of digital images of human faces, each subspace may be associated with a particular person for which a number of images have been collected, for instance under different lighting conditions, and the identification of these subspaces will aid face recognition \cite{McWilliams2011}. When the observations are deemed to be representative of $K$ distinct low-dimensional subspaces, the problem of {\it subspace clustering} then consists in the simultaneous estimation of these $K$ subspaces and the unobserved membership of data points to subspaces \cite{Vidal2011}. A formal definition of this problem as well as a brief survey of existing models and algorithms in provided in Section \ref{review}. 

In this work we introduce a novel approach to subspace clustering and present extensive comparisons to competing methods using both simulated and real high-dimensional genomic data sets. We exploit the fact that PCA provides the low-dimensional representation of the data that minimises the reconstruction error, and propose a criterion derived from the out-of-sample error as the building block for a subspace clustering algorithm.  An overview of the contributions presented here is in order.

First, for a single PCA model, we propose a computationally efficient approach to detect {\it influential observations}, namely data points exerting a larger effect on the estimated PCA parameters compared to other points. Our motivation for detecting influential observations is that, under the assumption that $K$ alternative low-dimensional representations of a data point might exist, a measure of influence will provide a useful metric for assigning data points to subspaces. A common way to quantify the influence of a data point consists of examining the changes in the model parameters when the parameters have been re-estimated after removal of that observation \cite{Belsley1980}. In the context of ordinary least squares (OLS), this is equivalent to computing the leave-one-out (LOO) estimate of the regression coefficients with respect to each observation. The LOO prediction error is then evaluated using the remaining observations, and influential observations are identified as those with a large LOO prediction error relative to other observations. Following up  previous work done in the context of partial least squares regression \cite{McWilliams2010a}, here we propose a closed-form expression to compute LOO errors under a PCA model, known at the PRESS statistic, which only requires a single PCA model to be fitted to the entire data set. Armed with this analytical expression for the PCA PRESS statistic, we propose a notion of {\it predictive influence} that an observation exerts on the PCA model, and elaborate on our previous work \cite{McWilliams2011}; intuitively, data points that are highly influential under a particular PCA model hay have been generated by a different model. This methodology is presented in Section \ref{sec_piPCA}.
 
Building on this notion of predictive influence, we develop a clustering algorithm that works by inferring the distinct low-dimensional spaces that are representative of each cluster. The optimality criterion we propose for driving the data partitioning process is such that total within-cluster predictive influence is minimised. The resulting algorithm, called {\it Predictive Subspace Clustering} (PSC) because of its direct dependence upon out-of-sample prediction errors, is an iterative one and is guaranteed to convergence to a local minimum. The convergence of the proposed algorithm has been studied, and we provide a detailed account. Although inferring the unknown number of clusters is a notoriously difficult problem, model selection can be somewhat naturally incorporated into the proposed subspace clustering framework by making use of the PCA PRESS statistic. Furthermore, motivated by genomic applications in which the detection of a small number of informative variables is important, we also discuss a variation of the PSC algorithm which provides sparse low-dimensional representations of the data in each cluster. Forcing sparse solutions within our clustering algorithm is accomplished by taking a penalised approach to PCA. These developments are presented in Section \ref{sec_ss}.

In Section \ref{results} we present extensive simulation results based on artificial data and discuss a number of situations where the proposed approach is expected to provide superior clustering results, compared to standard clustering methods as well as other subspace algorithms that have been proposed in the literature. Different scenarios are considered in which both the number of subspaces and their dimensions is allowed to vary, and where some dimensions may be totally uninformative for clustering purposes and only contribute to noise. The difficult model selection problem, that is the problem of learning both the number of clusters and the dimensionality of each subspace, is also discussed. 

In Section \ref{real_data} the performance of the proposed PSC algorithm is compared to other subspace clustering algorithms using six high-dimensional genomic data sets in which several thousands of gene expression measurements have been made on various biological samples. For these data sets, which are all publicly available, both the number of clusters and the cluster memberships can be considered known, and this information can be used to test and compare the performance of competing clustering methods. It has often been observed that only certain variables (i.e. genes) are important for determining the separation between clusters. It is therefore desirable to be able to identify and remove any uninformative variables. Our experimental results demonstrate that the assumptions underlying the proposed PSC algorithm seem plausible for real gene expression data sets on which we have obtained state-of-the-art clustering performance. 

\section{Subspace clustering: problem definition and existing methods} \label{review}


We assume to have observed $N$ data points, $\{\bm{x}_i\}_1^N$, where each $\bm{x}_i\R^{1\times P}$ and the dimension $P$ is usually very large. Each point is assumed to belong to one of $K$ non-overlapping clusters, $\{\mathcal{C}_k\}_1^K$. We further assume that the points in the $k^{th}$ cluster lie in a $R_k-$dimensional subspace, $\mathcal{S}_k$ where $R_k<<P$. As in \cite{Vidal2011}, we assume that each subspace $\mathcal{S}_k$ is defined in the following way 
\begin{equation} \label{eq_ssdef}
\mathcal{S}_k = \{  \bm{x}_i : \bm{x}_i = \bm{\mu}_{k} +  \bm{u}_{k,i} {\bm{V}_{k}}\tr \}  
\end{equation} 
with $i \in \mathcal{C}_k$ and $k=1,\ldots,K$, where $\bm{V}_{k}\R^{P\times R_k}$ is a basis for $\mathcal{S}_k$ whose columns are restricted to be mutually orthonormal. The point $\bm{u}_{k,i}\R^{R_k}$ is the low dimensional representation of $\bm{x}_i$ and $\bm{\mu}_{k}\R^{P}$ is an arbitrary point in $\mathcal{S}_k$, typically chosen to be $\bm{0}$.

When only one cluster exists (i.e. $K=1$), a subspace of this form can be estimated by fitting a single global PCA model. Alternatively, where $K>1$ and the assignment of points to clusters is known, each one of the $K$ subspaces of form \eqref{eq_ssdef} can be estimated by fitting a PCA model independently in each cluster. However, the cluster assignments are typically unknown, and the problem consists in the simultaneous partitioning the data into clusters and estimating cluster-specific subspaces. There are several fundamental difficulties associated with this problem: (a) identifying the true subspaces is dependent on recovering the true clusters and vice-versa; (b) subspaces can intersect at several locations which causes difficulties when attempting to assign points to subspaces at these intersections, and standard clustering techniques such as $K$-means may not be suitable; (c) the subspace parameters and the cluster assignments are dependent on both number of clusters and the dimensionality of their respective subspaces, which pose difficult estimation challenges.

Furthermore, in problems where $P$ is large, some of the variables may be uninformative for clustering. In some applications, there might also be an interest in selecting a specific subset of dimensions that are highly discriminative between clusters, and can be more easily interpreted. In such situations, we may be interested in carrying out {\it sparse} subspace clustering by adding some form of regularisation during the estimation of the eigenvectors in $\bm{V}$, such that each eigenvector contains a predetermined number of zero coefficients. To the best of our knowledge, the problem of variable selection in subspace clustering has not been widely studied until now. 

A variety of approaches have been proposed to solve the subspace clustering problem. Several methods are based on generalising the widely used $K$-means algorithm to $K$-subspaces \cite{Bradley2000,Wang2009}. These methods iteratively fit PCA models to the data and assign points to clusters until the PCA reconstruction error in each cluster is minimised. Although the approach based on minimising the within-cluster PCA reconstruction error is simple and has shown promising results, it is also prone to over-fitting. For instance, the data may be corrupted by noise or lie on the intersection between subspaces and so points within clusters may be geometrically far apart; we illustrate this problem using artificial data in Section \ref{results}. Furthermore, each subspace may have a different intrinsic dimensionality. The PCA reconstruction error decreases monotonically as the dimensionality increases, so points may be wrongly assigned to the cluster with the largest dimensionality. Such an approach therefore limits the number of dimensions to be the same in each cluster. Other approaches to subspace clustering have been taken based on mixtures of probabilistic PCA \cite{Tipping1999}. The recently proposed mixtures of common $t$-factor analysers (MCtFA) attempts to overcome the problems posed by over-fitting and potential outliers by assuming that the clusters share common latent factors which, instead of being normally distributed, follow a multivariate $t$-distribution \cite{Baek2011}. 

A different class of subspace clustering algorithms involves the computation of a measure of distance between each pair of points which captures the notion that points may lie on different subspaces. The distances are then used to construct an affinity matrix which is partitioned using standard spectral clustering techniques \cite{Luxburg2007}. There have been several successful approaches to defining such a distance measure. The method of Generalized PCA (GPCA) fits $K$ polynomials of varying order to the data and measures the distances between the gradient of the polynomials computed at each point \cite{YiMa2006}. Sparse subspace clustering (SSC) obtains a local representation of the subspace at each point as a sparse weighted sum of all other points \cite{Elhamifar2009}; this is obtained by minimising the reconstruction error subject to a constraint on the $\ell_1$ norm of the weights so that the few non-zero weights correspond to points lying on the same subspace. Spectral curvature clustering (SCC) constructs a multi-way distance between randomly sampled groups of points by measuring the volume of the of the simplex formed by each group \cite{Chen2008}. Points which lie on the same subspace will define a simplex of volume zero. Spectral local best flats (SLBF) estimates a local subspace for each point by fitting a PCA model to its nearest neighbours; it then computes pair-wise distances between the locally estimated subspaces corresponding to each point \cite{Zhang2010}. Although spectral methods have achieved state-of-the art results in some application domains such as clustering digital images, they come with their own limitations. Computing local subspaces for each point can be computationally intensive and requires additional tuneable parameters.

\section{Detecting influential observations in PCA} \label{sec_piPCA}

\subsection{Influential observations in ordinary least squares}

An influential observation is defined as one which, either individually or together with several other observations, has a demonstrably larger impact on the model estimates than is the case for most other observations \cite{Belsley1980}. Unlike outliers, influential observations are not necessarily apparent through simple visualisations of the data, and therefore a number of approaches have been proposed to detect them \cite{Imon2005}. Several popular methods rely on the computation of the leave-one-out (LOO) residual error \cite{Belsley1980,Meloun2001}. For instance, in the context of ordinary least squares (OLS), a common strategy consists in quantifying the effects that removing a single observation and re-fitting the model using the remaining $N-1$ observations has on the estimated regression coefficients \cite{Chatterjee1986}. 

When $N$ data points $\{ \bm{x}_i, y_i\}_{1}^N$ are available, where each $\bm{x}_i \R^P$ is a covariate and $y_i \R$ is the associated univariate response, the LOO error for the $i^{th}$ observation is defined as $\epsilon_{-i} = {y}_i - \bm{x}_i\bm{\beta}_{-i}$ where $\bm{\beta}_{-i}\R^{P\times 1}$ has been estimated using all but the $i^{th}$ observation. The sum of LOO errors is then as obtained as
\begin{equation} \label{ols_press}
J_{\text{OLS}} =  \frac{1}{N}\sum_{i=1}^N \lv  \epsilon_{-i} \rv^2.
\end{equation}

A naive computation of $J_{\text{OLS}}$ requires fitting $N$ OLS models, each one using $N-1$ observations. For each model fit, the inverse of sample the covariance matrix, $\bm{P}$, has to be estimated. In order to contain the number of computations needed to evaluate $J_{\text{OLS}}$, each LOO error can instead be found in closed-form after fitting a single regression model with all the $N$ data points. This is particularly beneficial when either $N$ or $P$ is large. Since each $\bm{\beta}_{-i}$ differs from $\bm{\beta}$ by only one pair of  observations, $\{\bm{x}_i,y_i\}$, a recursive formulation of $\bm{\beta}_{-i}$ that only depends on $\bm{\beta}$ is obtained by applying the Sherman-Morrison theorem \cite{Sherman1950}, as follows 
\begin{equation}
\bm{\beta}_{-i} =  \left( \bm{X}\tr \bm{X} - \bm{x}_i\tr\bm{x}_i \right)^{-1}\left(\bm{X}\tr \bm{y} - \bm{x}_{i}\tr y_{i}\right) 
=   \bm{\beta} - \frac{(y_i - \bm{x}_i \bm{\beta}) \bm{P} \bm{x}_i}{1 - \bm{x}_i\tr \bm{P} \bm{x}_i} \label{eq_ols_betai}.
\end{equation}
In this formulation, each $\bm{\beta}_{-i}$ is readily available as a function of ${\bm{\beta}}$, without the need to explicitly remove any observations, and without having to re-compute $N$ inverse covariance matrices. When this approach is taken, Eq. \eqref{ols_press} is known as the Predicted REsidual Sum of Squares (PRESS) statistic \cite{Belsley1980}. 

In the following Section we propose an analogous PRESS statistic for PCA and describe a methodology for detecting influential observations under a fitted PCA model. This approach will then be used to define an objective function for subspace clustering in Section \ref{sec_ss}. 

\subsection{An analytic PCA PRESS statistic} 

PCA is a ubiquitous method for dimensionality reduction when dealing with high-dimensional data. It relies on the assumption that the high-dimensional observations can be well approximated by a much lower-dimensional linear subspace which is found by estimating the best low-rank linear approximation of the original data \cite{Jolliffe}. One way of achieving this is by estimating a set of $R$ mutually orthonormal vectors $[\bm{v}^{(1)},\ldots,\bm{v}^{(R)}]$ which minimize the $\ell_2$ reconstruction error, defined as
\begin{equation}
\frac{1}{N}\sum_{i=1}^N \lv \bm{x}_i - \bm{x}_i \sum_{r=1}^{R} \bm{v}^{(r)}{\bm{v}^{(r)}}\tr  \rv^2. \label{eq_pca_recon}
\end{equation}

On defining a matrix $\bm{X}\R^{N\times P}$ with rows $\bm{x}_i$, which we assume to be mean-centred, the vectors which minimise \eqref{eq_pca_recon} are obtained by computing the singular value decomposition (SVD) of $\bm{X}$, given by
$
\bm{X}=\bm{U} \bm{\Lambda} \bm{V} \tr.
$
Here, $\bm{U}=[ \bm{u}^{(1)},...,\bm{u}^{(N)}] \R^{N \times N}$ and $\bm{V}=[ \bm{v}^{(1)},...,\bm{v}^{(P)}] \R^{P \times P}$ are orthonormal matrices whose columns are the left and right singular vectors of $\bm{X}$, respectively, and $\bm{\Lambda} =\text{diag}({\lambda^{(1)}},\ldots,{\lambda^{(N)}}) \R^{N \times P}$ is a diagonal matrix whose entries are the singular values of $\bm{X}$ in descending order. 

It can also be noted that, when taking the first principal component, the residual error can be written as a quadratic function of $\bm{v}^{(1)}$, 
\begin{equation}
\frac{1}{N}\sum_{i=1}^N\lv \bm{x}_i  - {d^{(1)}_i\bm{v}^{(1)}}\tr  \rv ^2,
\end{equation}
where $d^{(1)}_i = \bm{x}_i \bm{v}^{(1)}$. This formulation suggests that the eigenvector $\bm{v}^{(1)}$ can be rewritten in the form of a least squares estimator,
\begin{equation}
\bm{v}^{(1)} = \left( {\bm{d}^{(1)}} \tr \bm{d}^{(1)} \right)^{-1} \left(\bm{X}\tr \bm{d}^{(1)}\right) \label{eq_ls_pca},
\end{equation}
where we have defined the vector $\bm{d}^{(1)} = [d^{(1)}_1,\ldots, d^{(1)}_N ]\tr$; analogous expressions exist for the remaining eigenvectors, $\bm{v}^{(2)},\ldots,\bm{v}^{(R)}$. Clearly, each $\bm{d}^{(r)}$ depends on $\bm{v}^{(r)}$, and these eigenvectors are obtained as usual, using the SVD. This least squares interpretation will provide a starting point for deriving an efficient PRESS statistic for PCA.

In the context of PCA, the LOO error has often been used to drive model selection as well as detect influential observations \cite{Mertens1995,Bro2008}. When $R$ principal components are considered, this reconstruction error is given by
\begin{equation}
\bm{\epsilon}_{-i}^{(R)} = \bm{x}_i - \bm{x}_i\sum_{r=1}^{R} {\bm{v}^{(r)}_{-i} \bm{v}^{(r)}_{-i}}\tr,
\end{equation}
where each $\bm{v}_{-i}^{(r)}$ is the $r^{\text{th}}$ right singular vector of the SVD estimated using all but the $i^{th}$ observation of $\bm{X}$, and the sum of all LOO reconstruction errors is 
\begin{equation} \label{PCApress}
\tilde{J}^{(R)}_{\text{PCA}}= \frac{1}{N} \lv  \bm{\epsilon}_{-i}^{(R)} \rv^2 .
\end{equation}
The usual approach for the evaluation of $\tilde{J}^{(R)}_{\text{PCA}}$ requires $N$ SVD computations to be performed. Clearly, this approach is computationally expensive when either $N$ or $P$ is large, especially if \eqref{PCApress} has to be evaluated a number of times. 

We propose a novel closed-form approximation of \eqref{PCApress} which is computationally cheap and such that the approximation error is negligible for any practical purposes. The only assumption we make is that, for a sufficiently large sample size $N$, the error made by estimating the $R$ eigenvectors $\bm{v}^{(1)}, \bm{v}^{(2)}, \ldots, \bm{v}^{(R)}$ using the SVD of the reduced data matrix with $(N-1)$ rows, that is after removal of the $i^{\text{th}}$ observation, is sufficiently small, compared to the analogous estimation using the full data matrix. Under this assumption, we have that $\bm{v}_{-i}^{(r)}\approx \bm{v}^{(r)}$ and therefore $\bm{x}_i \bm{v}^{(r)}_{-i} \approx  \bm{x}_i \bm{v}^{(r)}$ for $i=1,\ldots,N$. We note here that a similar assumption has been made in other applications involving high-dimensional streaming data, where it is called {\it Projection Approximation Subspace Tracking} (PAST) \cite{Yang1996}. 

With only one principal component, this approximated PCA PRESS statistic can be written as
\begin{equation}
J^{(1)}_{\text{PCA}} =  \frac{1}{N}\sum_{i=1}^N \lv \bm{x}_i  - {d^{(1)}_i\bm{v}^{(1)}_{-i}}\tr \rv^2 \label{eq_loo}
\end{equation}
where $d^{(1)}_i = \bm{x}_i \bm{v}^{(1)}$, as before. Since $\bm{v}^{(1)}$ can be expressed as the solution to a least squares fit, as in \eqref{eq_ls_pca}, we can obtain $\bm{v}_{-i}$ in closed-form through least squares estimation. Using the relationships
$$
\sum_{j\neq i}^N {{d}^{(1)}_j}^2 = {\bm{d}^{(1)}} \tr \bm{d}^{(1)} - {d^{(1)}_i}^2, ~ \sum_{j\neq i}^N {\bm{x}_j d^{(1)}_j}=\bm{X} \bm{d}^{(1)} -\bm{x}_i d^{(1)}_i
$$
we arrive at an expression for $\bm{v}_{-i}$ by applying a single-observation, down-dating operation, as follows
\begin{equation}
\bm{v}^{(1)}_{-i} = \left( {\bm{d}^{(1)}} \tr \bm{d}^{(1)} - {d^{(1)}_i}^2 \right)^{-1} \left(\bm{X}\tr \bm{d}^{(1)} -\bm{x}_i\tr d^{(1)}_i\right) \label{eq_vi1}.
\end{equation}

The advantage of this reformulation is that Eq. \eqref{eq_vi1} can be evaluated using the same recursive form given by Eq. \eqref{eq_ols_betai}. Using the same recursive approach, we then obtain an expression for $\bm{v}^{(1)}_{-i}$ in terms of the original eigenvector $\bm{v}^{(1)}$,
\begin{equation}
\bm{v}^{(1)}_{-i} = \bm{v} ^{(1)}- \frac{ \left( \bm{x}_i\tr - d^{(1)}_i \bm{v}^{(1)} \right)D^{(1)} d^{(1)}_i}{1 - h^{(1)}_i},
\label{eq_matinv}
\end{equation}
where $h^{(1)}_i = d^{(1)}_i D^{(1)} d^{(1)}_i $, $D^{(1)}=\left({\bm{d}^{(1)}}\tr\bm{d}^{(1)}\right)^{-1}$, and $\bm{v}^{(1)}$ has been obtained by computing the SVD of the complete data matrix $\bm{X}$ in the usual way. Using this recursive expression, the $i^{\text{th}}$ LOO reconstruction error is obtained in closed-form   
\begin{equation}
\bm{e}^{(1)}_{-i} = \bm{x}_i - d^{(1)}_i\left( \bm{v}^{(1)} - \frac{ \left( \bm{x}_i\tr - d^{(1)}_i \bm{v}^{(1)} \right)D^{(1)} d^{(1)}_i}{1 - h^{(1)}_i} \right)\tr.
\end{equation}
By substitution, using the fact that the the $i^{th}$ reconstruction error is $\bm{e}^{(r)}_i=\bm{x}_i - {\bm{x}_i\bm{v}^{(r)}\bm{v}^{(r)}}\tr$, we then obtain
\begin{equation}
\bm{e}^{(1)}_{-i} = \bm{e}^{(1)}_i \left( 1 +  \frac{h^{(1)}}{{1-h^{(1)}}}\right) \frac{\bm{e}^{(1)}_i}{1 - h^{(1)}_i},
\end{equation}
which can be computed in closed form, without the need for explicit LOO steps. This construction can be easily extended to the case of $R>1$ principal components. When $R=2$, we have
\begin{equation*}
\bm{e}^{(2)}_{-i}
 = \bm{x}_i  - {d^{(1)}_i\bm{v}^{(1)}_{-i}}\tr - {d^{(2)}_i\bm{v}^{(2)}_{-i}}\tr = \frac{\bm{e}^{(1)}_i}{1 - h^{(1)}_i} - {d^{(2)}_i\bm{v}^{(2)}_{-i}}\tr.
\end{equation*}
Adding and subtracting $\bm{x}_i$ from both sides we obtain
\begin{equation*}
\bm{e}^{(2)}_{-i}
 = \frac{\bm{e}^{(1)}_i}{1 - h^{(1)}_i} + \bm{x}_i - {d^{(2)}_i\bm{v}^{(2)}_{-i}}\tr - \bm{x}_i = \sum_{r=1}^{2} \frac{\bm{e}^{(r)}_i}{1 - h^{(r)}_i} - \bm{x}_i,
\end{equation*}
where $\bm{e}^{(r)}_i = \bm{x}_i - \bm{x}_i\bm{v}^{(r)}{\bm{v}^{(r)}}\tr $. Therefore, with $R$ principal components, the approximated PCA PRESS statistic is \begin{equation}
J^{(R)}_{\text{PCA}} = \frac{1}{N}\sum_{i=1}^{N} \lv  \sum_{r=1}^{R} \frac{\bm{e}^{(r)}_i}{1 - h^{(r)}_i} - (R-1)\bm{x}_i \rv^2.  \label{pca_press}
\end{equation}

\begin{figure}[hpt!]
\begin{center}
\includegraphics[width=5in]{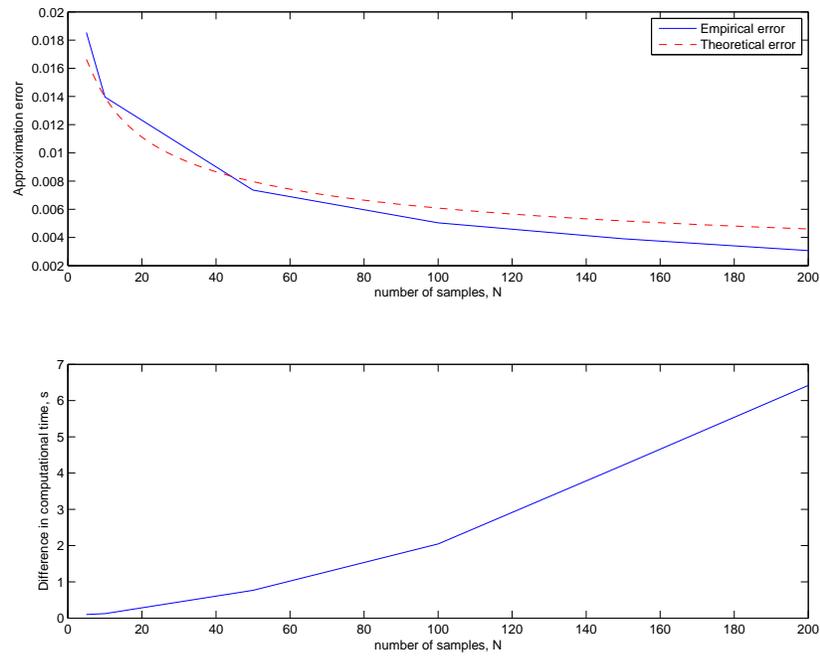}%
\caption{The top plot shows the mean squared approximation error between the leave-one-out cross validation $J_{PCA}$ and our analytic PRESS statistic $\tilde{J}_{PCA}$ as a function of the number of samples, $N$. We report on averages over $100$ Monte Carlo simulations. It can be seen that the empirical approximation error scales according to the theoretical error, which is shown as a dashed line. The bottom plot shows that the difference in computational time between $J_{PCA}$ and the $\tilde{J}_{PCA}$ increases super-linearly with $N$.\label{fig_press_approx}}
\end{center}
\end{figure}

This expression only depends on quantities estimated using a single PCA model fit, and can be computed cheaply. The approximated $J^{(R)}_{\text{PCA}}$ is found to be close to the true PCA PRESS $\tilde J^{(R)}_{\text{PCA}}$ under the assumption that the error made by estimating the $R$ principal components by $N-1$ instead of $N$ observations is small. In related work, we have previously shown that the approximation error depends on the number of observations as $O(\sqrt{\log N}/N)$ \cite{McWilliams2010a}. This result can also be checked by simulation. Figure \ref{fig_press_approx} shows the mean squared approximation error as a function of $N$ using data simulated under a PCA model with $P=500$ variables. It can be seen that the decrease in the approximation error follows the theoretical error (plotted as a dashed line) and the difference in computational time increases super-linearly.


\subsection{A measure of predictive influence in PCA} \label{influence}

In this section we make use of the closed-form PCA PRESS statistic of Eq. \eqref{pca_press}, and propose an analytic influence measure for the detection of influential observations under a fitted PCA model. Specifically, we introduce the notion of {\it predictive influence} of an observation $\bm{x}_i$ under a given PCA model which quantifies how much influence $\bm{x}_i$ exerts on the LOO prediction error of the model. By making explicit use of the out-of-sample reconstruction error, this influence measure is particularly robust against over fitting compared to the in-sample reconstruction error that is minimised by PCA, as in Eq. \eqref{eq_pca_recon}.  

As before, we assume that a PCA model has been fit to the data, and $\bm{V}=[ \bm{v}^{(1)},...,\bm{v}^{(P)}] \R^{P \times P}$ is the orthonormal matrix whose columns are the right singular vectors of $\bm{X}$. Analytically, the predictive influence of observation $\bm{x}_i$, denoted here $\bm{\pi}(\bm{x}_i; \bm{V})$ is defined as the gradient of the PCA PRESS given in Eq. \eqref{pca_press} with respect to observation $\bm{x}_i$, that is
\begin{equation}
\bm{\pi}(\bm{x}_i; \bm{V})  =  \frac{\partial J^{(R)}_{\text{PCA}}}{\partial \bm{x}_i} \label{pred_mes}.
\end{equation}

Using the results from the previous section, an analytical expression for the gradient in Eq. \eqref{pred_mes} can be derived, and is found to be 
\begin{equation} \label{final_press}
\bm{\pi}(\bm{x}_i; \bm{V})  = \bm{e}^{(R)}_{-i}
  \left( \sum_{r=1}^{R} \frac{ \left(\bm{I}_p - \bm{v}^{(r)}{\bm{v}^{(r)}}\tr\right) }{\left(1-h^{(r)}_i\right)} - (R-1) \right).
\end{equation}
The full details of this derivation are given in Appendix \ref{sec_appA}. It can be seen that the predictive influence of a point, $\bm{x}_i$ under a PCA model has a closed form which is related to the leave-one-out error of that point, $\bm{e}_{-i}$.

It was observed in \cite{Cook1986} that single deletion methods for identifying influence, such as the PRESS, are only sensitive to large errors. Instead they propose observing the change in the error as each observation is weighted between 0 (equivalent to LOOCV) and 1 (equivalent to the standard residual) and then computing the derivative of the residual with respect to the weight. On the other hand, our proposed predictive influence measures the sensitivity of the prediction error in response to an incremental change in $\bm{x}_i$. The rate of change of the PCA PRESS at this point is given by the magnitude of the predictive influence vector, $\lv \bm{\pi}(\bm{x}_i; \bm{V})\rv^2$. If the magnitude of the predictive influence is large, this implies a small change in the observation will result in a large change in the prediction error relative to other points. In this case, removing such a point from the model would cause a large improvement in the prediction error. We can then identify the most influential observations as those for which the increase in the PRESS is larger relative to other observations. Since we take the gradient of the PRESS with respect to the observations, we arrive at a quantity which is more sensitive to individual influential observations than the original PRESS function. In this work, the predictive influence measure has been used in the context of subspace clustering, which is introduced next.  


\section{Predictive subspace clustering} \label{sec_ss}

\subsection{The PSC algorithm} \label{sec_PSC}

The proposed algorithm for subspace clustering relies on the following observation. If the cluster assignments $\{\mathcal{C}_k\}_1^K$ were known and a separate PCA model was fit to the data in each cluster, then the predictive influence of a point $\bm{x}_i$ belonging to cluster $\mathcal{C}_k$ would be smaller when evaluated using the correct PCA model for that cluster, than using any of the remaining $K-1$ PCA models. In this respect, the predictive influence provides a metric that can be used to drive the clustering process. Since our clustering algorithm is based on a measure of predictive ability of the estimated subspaces, we call it {\it Predictive Subspace Clustering} (PSC).

The objective of the clustering algorithm is to partition the $N$ observations into one of $K$ non-overlapping clusters such that each cluster contains exactly $N_k$ observations and $\sum_{k=1}^K N_k=N$ and where the points in each cluster lie on a subspace of the form described in Eq. \eqref{eq_ssdef}.   In our proposal, this will be accomplished by searching for the cluster allocations and PCA model parameters that minimise the following objective function,
\begin{equation} \label{objective}
C(\bm{V}_1, \ldots, \bm{V}_k, \mathcal{C}_1, \ldots, \mathcal{C}_K) = \sum_{k=1}^{K} \sum_{i \in \mathcal{C}_k} \lv \bm{\pi}_{k}(\bm{x}_i; \bm{V}_k)\rv^2  ,
\end{equation}
where $\bm{\pi}_{k}(\bm{x}_i; \bm{V}_k)$ is the predictive influence of a point $\bm{x}_i$ under the $k^{th}$ PCA model. Since the cluster allocation and PCA model parameters depend on each other, there is no analytic solution to this problem and we must resort to an iterative procedure. 

This problem can be approached by considering the two related optimisation tasks. At a generic iteration $\tau$, given the $K$ subspaces with parameters $\bm{V}_1,\ldots,\bm{V}_K$ which were estimated in the previous iteration $\tau-1$, we search for the cluster assignments minimising \eqref{objective},
\begin{equation} \label{step1}
	\min_{ \mathcal{C}_1,...,\mathcal{C}_K } C(\bm{V}_1^{(\tau-1)}, \ldots, \bm{V}_K^{(\tau-1)}, \mathcal{C}_1, \ldots, \mathcal{C}_K).
\end{equation}
Using the predictive influences $\bm{\pi}_{k}^{(\tau-1)}(\bm{x}_i; \bm{V}_k^{(\tau-1)})$, which were computed at the end of the previous iteration for all $i=1,\ldots,n$ and all $k=1,\ldots, K$, a solution for \eqref{step1} is found by assigning each point $\bm{x}_i$ to the cluster for which the magnitude predictive influence is smallest, that is
\begin{equation}
	\mathcal{C}_k^{(\tau)} \leftarrow \left\{i: \min_k \lv \bm{\pi}_{k}^{(\tau-1)}(\bm{x}_i; \bm{V}_K^{(\tau-1)}) \rv^2 \right\}. \label{eq_assignrule}
\end{equation}

Then, in the second step, given the new cluster assignments $\{ \mathcal{C}_k^{(\tau)} \}_1^K$, we re-estimate the parameters of the $K$ subspaces minimising \eqref{objective} by solving
\begin{equation} \label{step2}
\min_{ \bm{V}_1,\ldots, \bm{V}_K } C(\bm{V}_1, \ldots, \bm{V}_K, \mathcal{C}_1^{(\tau)}, \ldots, \mathcal{C}_K^{(\tau)}). 
\end{equation}

Using the current cluster assignments, a solution for Eq. \eqref{step2} is found by re-fitting all $K$ PCA models because, from Eq. \eqref{final_press}, minimising the PCA construction error is also seen to minimise the predictive influences. Once all the parameters $\{\bm{V}^{(\tau)}_k\}_1^K$ have been re-estimated, the predictive influence measures   $\bm{\pi}_{k}^{(\tau)}(\bm{x}_i; \bm{V})$ for all $k=1,2,\ldots,K$ and $i=1,2, \ldots,N$, are updated for the subsequent iteration.

The algorithm is initialised by generating an initial random partitioning, $\{ \mathcal{C}_k^{(0)} \}_1^K$, which is used to estimate the $K$ initial PCA models, find the initial parameters $\{\bm{V}^{(0)}_k\}_1^K$, and compute the corresponding predictive influences, $\bm{\pi}_{k}^{(0)}(\bm{x}_i;\bm{V})$ for $k=1,\ldots,K$ and $i = 1,\ldots,N $. At each iteration $\tau=1, 2, \ldots$, the algorithm alternates between the two steps described above until convergence is reached and the objective function can no further be reduced. 

\subsection{Convergence of the PSC algorithm}

In this section we demonstrate that, for any initial configuration, the PSC algorithm is guaranteed to converge to a local minimum of the objective function \eqref{objective}. In order to keep the notation simple, and without any loss of generality, we only discuss the case of the first principal component. Furthermore, we denote $\mathcal{S}^{(\tau)}=\{\bm{v}^{(\tau)}_1, \ldots, \bm{v}^{(\tau)}_K\}$ the set of all PCA parameters at a generic iteration of the algorithm, $\tau$; we also use $\mathcal{C}^{(\tau)} = \{\mathcal{C}^{(\tau)}_1, \ldots, \mathcal{C}^{(\tau)}_K\}$ to denote the set of clustering assignments at the same iteration. With this notation in place, in order to show the algorithm converges we must demonstrate that, at each iteration $\tau$, the following inequalities are satisfied:

\begin{itemize}
\item[(a)] $C(\mathcal{S}^{(\tau-1)},\mathcal{C}^{(\tau)}) \leq C(\mathcal{S}^{(\tau-1)},\mathcal{C}^{(\tau-1)})$, after the first step, \eqref{step1};
\item[(b)] $C(\mathcal{S}^{(\tau)},\mathcal{C}^{(\tau)}) \leq C(\mathcal{S}^{(\tau-1)},\mathcal{C}^{(\tau)})$, after the second step, \eqref{step2}.
\end{itemize}


Checking that the first inequality holds after the first step of the algorithm is straightforward because, by definition, the cluster assignments $\mathcal{C}^{(\tau)}$ obtained by the assignment rule \eqref{eq_assignrule} directly minimise the objective function when the PCA parameters are held fixed. The second inequality, however, requires a more elaborated argument. 

Assuming that the clustering memberships are given and held fixed, we first note that solving \eqref{step2} is equivalent to solving the following problem
\begin{align}
\label{eq_rpca_opt}
& \min_{\bm{v}}  \sum_{i=1}^{N} \lv \bm{\pi}(\bm{x}_i; \bm{v}) \rv^2, \\ & \text{subject to } ~ \lv \bm{v} \rv =1, \nonumber
\end{align}
separately for each cluster.  The following lemma provides an alternative formulation of this optimisation problem. 

\begin{lem_pca}
Solving the minimisation problem \eqref{eq_rpca_opt} is equivalent to solving the following maximisation problem
\begin{align}
& \max_{\bm{v}}
\bm{v} \tr \bm{X} \tr {\bm{\Xi}}^{-2} \bm{X} \bm{v} , 
\label{eq_pca_max_pi} \\
& \text{subject to } ~ \lv \bm{v} \rv =1 .\nonumber
\end{align}
where $\bm{\Xi}\R^{N\times N}$ is a diagonal matrix with diagonal elements ${\Xi}_{i} = (1-h_i)^2$.
\label{lem_pca_maxprob}
\end{lem_pca}

The proof of this lemma is provided in Appendix \ref{proof_pca_maxprob}. The maximisation in \eqref{eq_pca_max_pi} can be recognised as an eigenproblem where each observation has been weighted by a function of its leverage under the PCA model. We denote the optimal parameters that provide a solution to this problem as $\mathcal{S}^{*}=\{ \bm{v}^{*}_k\}_1^K$. If such solution was available, it would satisfy inequality (b) above, so that
$$
C(\mathcal{S}^*,\mathcal{C}^{(\tau)}) \leq C(\mathcal{S}^{(\tau-1)},\mathcal{C}^{(\tau)}).
$$

The solution $\mathcal{S}^{*}$ depends on the diagonal element of $\bm{\Xi}$, which in turn depend on the PCA parameters and are non-linear function of $\bm{v}$, and therefore a closed-form solution to \eqref{eq_pca_max_pi} cannot be found simply by eigendecomposition of $\bm{X} \tr \bm{\Xi}^{-2} \bm{X}$. However, using this formulation, we are able to prove that the PCA parameters at iteration $\tau$ are closer to the optimal solution $\mathcal{S}^{*}$ than the parameters from the previous iteration, i.e.  $\mathcal{S}^{(\tau-1)}$.
 
\begin{lem_pca}
For a single cluster $k$, we define the error between the the optimal parameters ${\bm{v}_k^{*}}$ obtained by solving \eqref{eq_pca_max_pi} directly and the PCA parameters from the previous iteration, ${\bm{v}^{(\tau-1)}_{k}}$, as  
$$
  {E}(\mathcal{S}^*,\mathcal{S}^{(\tau-1)}) = \sum_{i\in\mathcal{C}_k^{(\tau)}}{\bm{v}_k^{*}}\tr \bm{x}_i\tr {\Xi^{-2}_{k,i}}  \bm{x}_i{\bm{v}_k^{*}} - \sum_{i\in\mathcal{C}_k^{(\tau)}} {\bm{v}^{(\tau-1)}_{k}}\tr \bm{x}_i\tr \bm{x}_i {\bm{v}^{(\tau-1)}_k}.
$$
Analogously, the error between the optimal parameters and the PCA parameters obtained at the current iteration $\tau$ is defined as  
$$
 {E}(\mathcal{S}^*,\mathcal{S}^{(\tau)}) = \sum_{i\in\mathcal{C}_k^{(\tau)}} {\bm{v}_k^{*}}\tr \bm{x}_i\tr {\Xi^{-2}_{k,i}} \bm{x}_i\bm{v}^{*}_{k} - \sum_{i\in\mathcal{C}_k^{(\tau)}} {\bm{v}^{(\tau)}_{k}}\tr \bm{x}_i\tr \bm{x}_i {\bm{v}^{(\tau)}_{k}}.
$$
These two error terms satisfy the inequality
\begin{align}
   {E}(\mathcal{S}^*,\mathcal{S}^{(\tau)}) \leq   {E}(\mathcal{S}^*,\mathcal{S}^{(\tau-1)}) .
\end{align}
\label{lem_pca_svd}
\end{lem_pca}
The proof of this lemma is provided in Appendix \ref{proof_pca_svd}. We are now ready to formulate the main result stating the convergence of the PSC algorithm.

\begin{thm_pca}
Starting with any cluster configuration, $\{ \mathcal{C}^{(\tau-1)}_k\}_1^K$, at each iteration of the PSC algorithm the objectives function \eqref{objective} is never decreased, and the algorithm converges to a local minimum.
\label{thm_pca_conv}
\end{thm_pca}

\begin{proof}
The proof of this theorem simply follows from the observations made above, and the two lemmas, which show that the inequalities (a) and (b) are satisfied at each iteration.

\end{proof}%


\subsection{Sparse subspace clustering}

As mentioned in Section \ref{introduction}, there may be many unimportant variables that only contribute to noise and therefore could be discarded. Moreover, in some applications, there might be the need to select only a handful of informative variables that best represent the subspace in each cluster.  Since the principal components involve linear combinations of all $P$ variables in $\bm{X}$, we present a variation of the PSC algorithm that build on a  sparse PCA model estimation procedure. 

Following the strategy described by \cite{Shen2008}, sparse loading vectors can be obtained by imposing an $\ell_1$ penalty on the PCA objective function \eqref{eq_pca_recon}. When $R=1$, this new optimisation problem become
\begin{align}
& \min_{{\bm{u}},{\bm{v}}}\lv \bm{X}-\bm{u}\bm{v}\tr \rv_F^{2}+ \gamma \lv {\bm{v}}\rv_1 \label{spca} \\
& \text{subject to } \lv \bm{u} \rv = 1 .\nonumber
\end{align}
where $\lv \bm{A} \rv_F^2=\text{Tr}(\bm{A}\tr \bm{A})$ denotes the squared Frobenius norm.
This problem can be solved by first obtaining $\bm{u}=\bm{u}^{(1)}$ and $\bm{v}=\sigma^{(1)}\bm{v}^{(1)}$ where $\bm{u}^{(1)}$,$\bm{v}^{(1)}$ and $\sigma^{(1)}$ are the first left and right singular vectors and corresponding singular value computed from the SVD of $\bm{X}$. A sparse solution is obtained by applying the following iterative soft thresholding procedure to the elements of $\bm{v}$:
\begin{align}
\label{it_penalty}{\bm{v}}  = & \text{sgn}\left(\bm{X}\tr {\bm{u}}\right)\left(\left| \bm{X}\tr {\bm{u}}\right|-\gamma\right)_{+}\\
\label{it_penalty2}
{\bm{u}}  = & \frac{\bm{X}{\bm{v}}}{\lv \bm{X}{\bm{v}}\rv}.
\end{align}

The updates \eqref{it_penalty} and \eqref{it_penalty2} are applied iteratively until the change in $\bm{v}$ between iterations falls below some threshold. Subsequent sparse loadings can be found by deflating the data matrix and repeating the above steps as in \cite{Shen2008}. The complexity of this procedure when $R$ sparse principal components are required is $O(NPR)$, which keeps the computational burden relatively low.

A penalised version of the PSC algorithms is obtained by modifying the second step of the algorithm so that \eqref{step2} is replaced accordingly; in the case of the first principal component, this amounts to solving 
\begin{align}
& \min_{\{\bm{v}_{1}, \ldots, \bm{v}_{K} \}} C (\bm{v}_1, \ldots, \bm{v}_k, \mathcal{C}_1, \ldots, \mathcal{C}_K) \\
& \text{subject to } || \bm{v}_{k} ||_1 \leq \gamma_k \text{ for } k=1,\ldots,K  \nonumber
\end{align}
where the parameter which controls the level of sparsity, $\gamma_k$, can in principle be different for each cluster. It can be seen that there are $K$ inequality constraints, one for each of the subspaces.
This sparse version of the PSC algorithm is detailed in Algorithm \ref{alg_PSC}. Clearly, when the sparsity parameters are taken to be zero, we obtain the the unpenalised version of the algorithm and no variable selection is performed.
\begin{figure}

\begin{algorithm}[H]
\caption{The penalised PSC algorithm.}
\KwIn{$X$, $K$, $R_k$, $\gamma_k$, $\mathcal{C}_k$, for $k=1, \ldots, K$}
\KwOut{$\{ \mathcal{C}_1,\ldots,\mathcal{C}_K \}$, $\{ \bm{V}_1,\ldots,\bm{V}_K$\}}

\While{not converged}{
\For{$k=1,\ldots,K$}{
\tcp{Compute the sparse subspace parameters}
$\bm{X}_k\leftarrow \{\bm{x}_i\} \in \mathcal{C}_k$ \;
$[\bm{u}^{(1)},\sigma^{(1)},\bm{v}^{(1)}] \leftarrow {svd}(\bm{X}_k,R)$\;
\For{$r=1,\ldots,R_k$}
{
\While{change in $\bm{v}^{(r)}>tol$}{
${\bm{v}^{(r)}}  \leftarrow  \text{sgn}\left(\bm{X}\tr {\bm{u}^{(r)}}\right)\left(\left| \bm{X}\tr {\bm{u}^{(r)}}\right|-\gamma\right)_{+}$ \;
${\bm{u}^{(r)}}  \leftarrow  \frac{\bm{X}{\bm{v}^{(r)}}}{\lv \bm{X}{\bm{v}^{(r)}}\rv}$ \;
}
$\bm{X}_k \leftarrow \bm{X}_k - {\bm{u}^{(r)}\bm{v}^{(r)}}\tr$
}
$\bm{V}_k \leftarrow [\bm{v}^{(1)},\ldots,\bm{v}^{(R)}]$\;
Compute PRESS, $J_k$\;
\For{$i=1,\ldots,N$}{
Compute predictive influence, $\bm{\pi}_k(\bm{x}_i;\bm{V})$\; 
}
}
\tcp{Assign points to clusters}
\For{$i=1,\ldots,N$}{
	$\mathcal{C}_k^{(\tau)} \leftarrow \left\{i: \min_k \lv \bm{\pi}_{k}^{(\tau-1)} (\bm{x}_i;\bm{V}) \rv^2 \right\}.$
}
\tcp{Check for convergence}
\If{$\sum_{k=1}^K J_k^{(\tau)} -  J_k^{(\tau-1)} > tol$}{converged\;}
$\tau \leftarrow \tau+1$\;
}
\label{alg_PSC}
\end{algorithm}
\end{figure}

\subsection{Model selection}
\label{sec_modsel}

Model selection in subspace clustering consists of learning both the number of clusters and the dimensionality of each subspace. Each one of these two problems is particularly challenging on its own. Since the PSC algorithm is a non-probabilistic one, traditional model selection techniques for learning the number of clusters such as the Bayesian Information Criterion and related methods do not apply. Moreover, it has already been noted that such methods, when can be used, do not always select optimal models that maximise clustering accuracy \cite{Baek2011}. 

In the subspace clustering literature, the problem of model selection is still in its infancy. A method called {\it second order difference} (SOD) has recently been proposed to determine the correct number of clusters \cite{Zhang2010}. The SOD method can be seen as an extension of the gap statistic originally proposed by \cite{Tibshirani2002}, and works as follows. For each value of $K$ up a maximum number of clusters, $K_{max}$, the cluster assignments and corresponding subspaces as in \eqref{eq_ssdef} are estimated. For each value of $K$, we compute the distance between points and subspaces within each cluster, as follows
$$
W_K = \frac{1}{K}\sum_{k=1}^{K} \frac{1}{N_k}\sum_{i\in \mathcal{C}_k} \lv \bm{x}_i - \bm{x}_i \bm{V}_k \bm{V}_k \rv^2.
$$
The second derivative of the within-cluster residual error with respect to the number of clusters $K$ is approximated by
$$
SOD(\log W_K)=\log W_{K - 1} +\log W_{K+1} - 2\log W_K,
$$
and this quantity is used as a search criterion. The optimal number of clusters, $K^{*}$ is then chosen to be the value that maximises this criterion, 
$$
K^{*} =\max_K SOD (\log W_K).
$$
This optimal choice corresponds to the point beyond which increasing the number of clusters has less effect on the within-cluster residual error.

Apart from the SOD method, our proposed PCA PRESS statistic \eqref{pca_press} also naturally provides an alternative criterion for model selection. As with SOD, we initially assume that the dimensionality of each subspace is known, and is the same for all subspaces, so that $R_k=R$ for all $k$. Using a varying number of clusters up to a pre-determined maximum, $K_{max}$, we run the PSC algorithm and, for each $K$, record the corresponding value of the PCA PRESS statistic, that is $J^{(R)}_{\text{PCA}}(K)$. The optimal number of cluster is found as 
$$
K^{*} =\min_K J^{(R)}_{\text{PCA}}(K).
$$
This approach is somewhat similar to SOD, however instead of measuring the within-cluster PCA residual error, which may suffer from over-fitting, we use the within cluster LOO cross-validation error, which negates the requirement to compute the second derivative with respect to $K$. The PCA PRESS is robust against over fitting and has been found to work generally well in our experiments presented in the next section.

The problem of learning the subspace dimensions $\{R_k\}_1^K$ has not been well studied in the subspace clustering literature, and is very much an open question. However, the PRESS has often been used to perform model selection in standard PCA, and seems to be well suited for this problem. With a given fixed $K$, at each generic iteration $\tau$ of the PSC algorithm, we evaluate the PCA PRESS using all values of $R_k^{(\tau)}$ from one to a pre-determined maximum, $R_{max}$, and then select the set $\{ R_k^{(\tau)} \}_1^K$ of subspace dimensions that minimises the PCA PRESS at that iteration. Although computationally more expensive, this strategy has often been found successful in learning these parameters. Some experimental results are reported in the next section. 

Finally, an important feature of the PSC algorithm is its ability to estimate sparse PCA models. In order to obtain sparse models, we need to specify additional parameters $\{ \gamma_k\}_1^K$ which control the number of variables to be retained within each cluster. This introduces further complications to the issue of model selection. In the context of sparse predictive modelling, it has often been observed that prediction-based methods such as the PRESS do not perform well for selecting the optimal sparsity parameter. Recently, subset resampling methods such as {\it stability selection} have shown promising results for accurately selecting regularisation parameters \cite{Meinshausen2010}. However, implementing such data resampling schemes within the PSC algorithm would be computationally prohibitive. 

\section{Simulation experiments} \label{results}

We start by presenting a number of simple simulation studies in low dimensions to illustrate the type of clusters that can be detected by the PSC algorithm, and compare the performance of the proposed algorithm to existing methods. Clusters of $100$ data points were generated and, within each cluster, the points were distributed uniformly on a one, two and three-dimensional linear subspace embedded in a three-dimensional space. To define each subspace, we generated a set of $R_k$ orthonormal basis vectors, each of dimension $P=3$, where each element was sampled from a standard normal distribution. For each cluster we then sampled $100$ $R_k$-dimensional points from a uniform distribution which were then projected onto its corresponding subspace.

\begin{figure}
\centering
\subfloat[Two lines]{\includegraphics[width=5.5in]{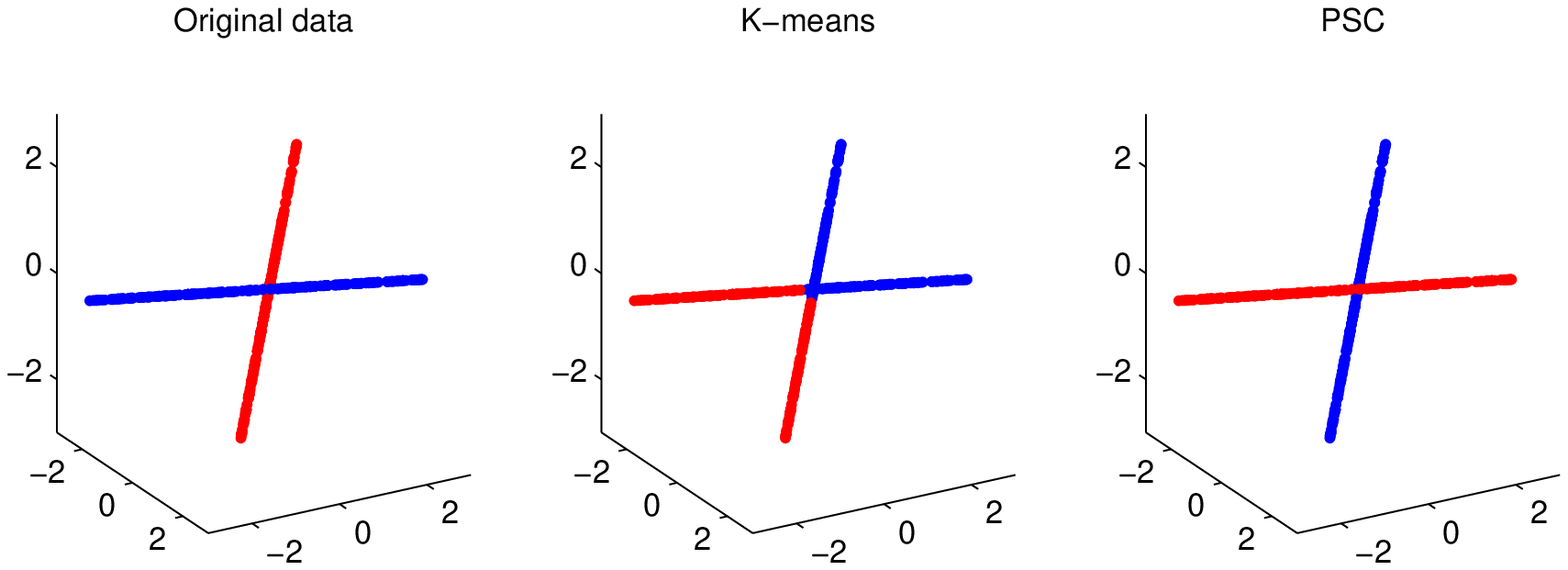}%
\label{fig_2lines}}
\hfil
\subfloat[Two subspaces: Line and plane]{\includegraphics[width=5.5in]{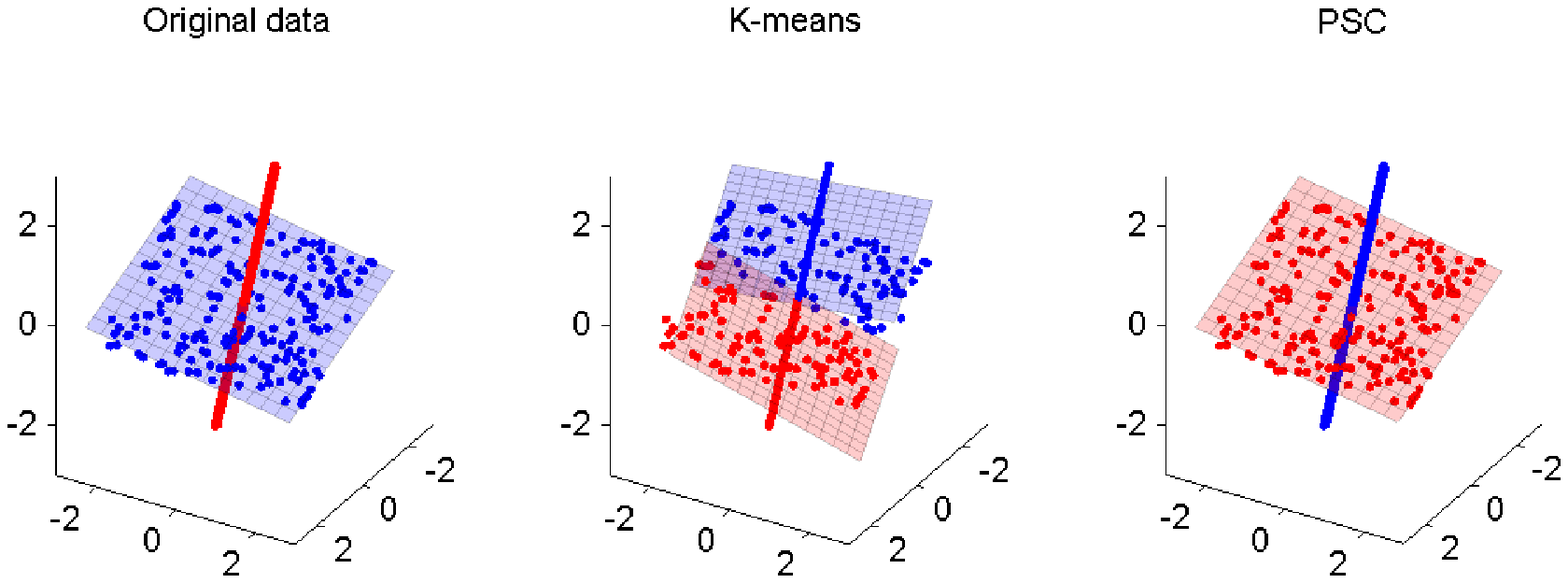}%
\label{fig_lineplane}} \\
\subfloat[Two planes]{\includegraphics[width=5.5in]{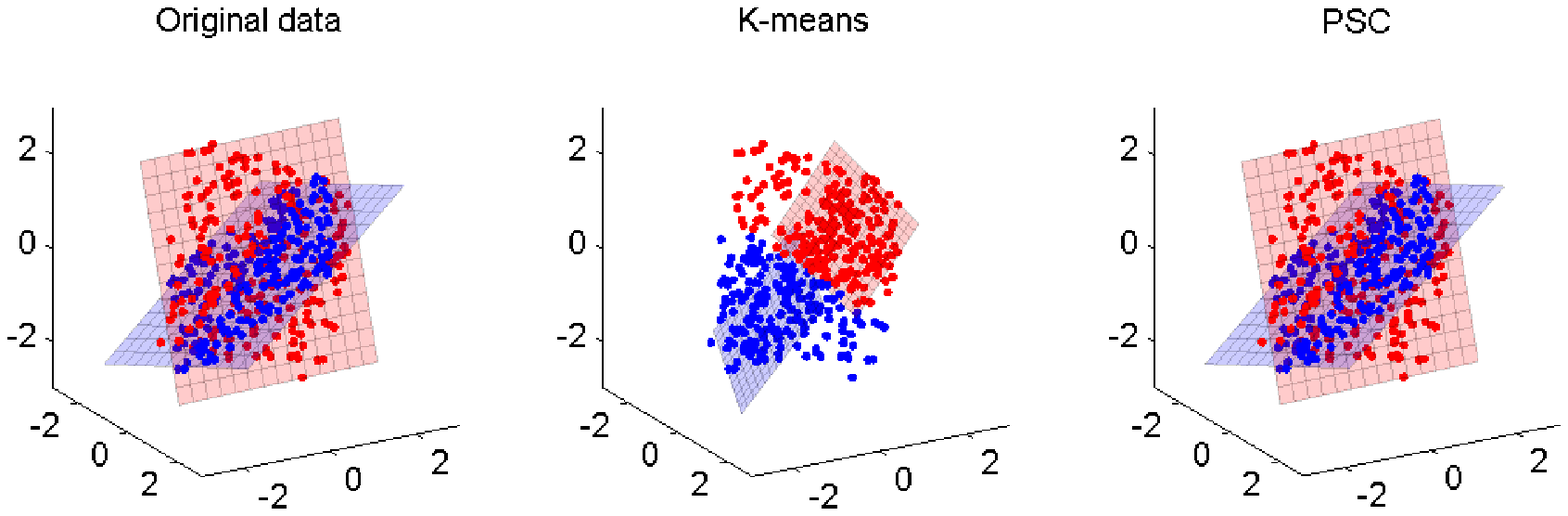}%
\label{fig_2planes}}
\hfil
\subfloat[Three subspaces: Line, plane and sphere]{\includegraphics[width=5.5in]{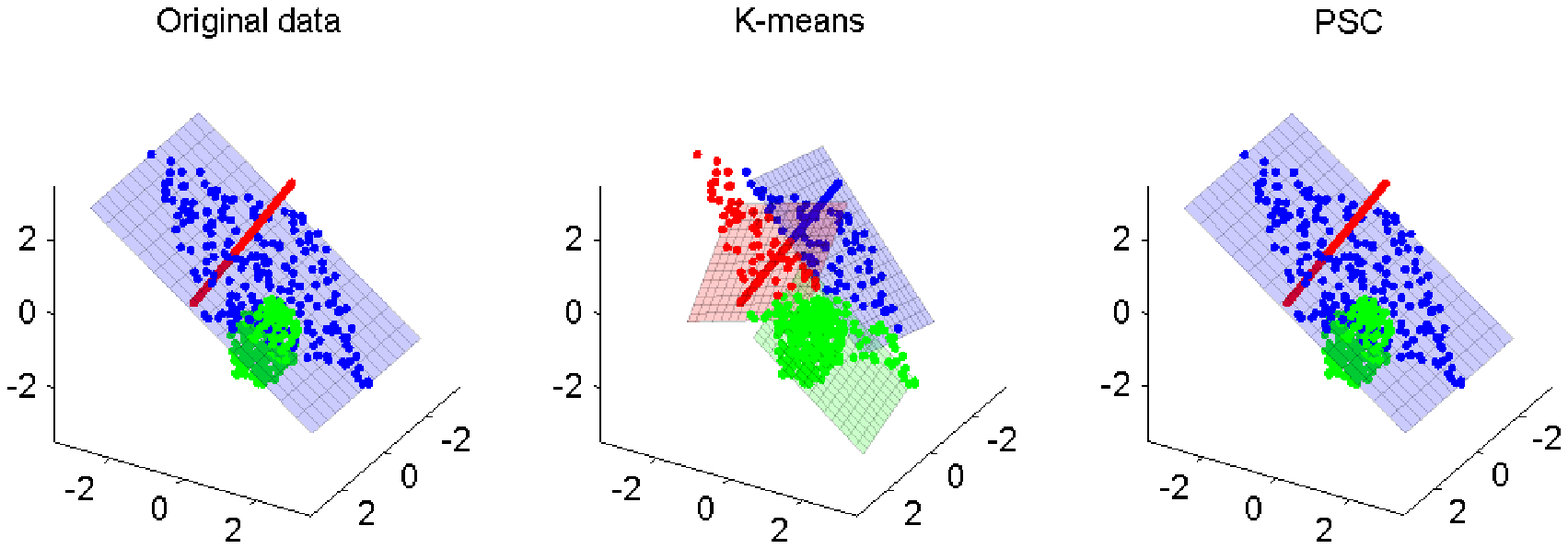}%
\label{fig_lineplaneball}}
\caption{Four illustrations of data clusters existing in subspaces of $\mathbb{R}^3$ and the clusters recovered by two algorithms, K-means and the proposed PSC algorithm. In these examples PSC recovers the true cluster assignments and estimates the subspaces correctly.}
\label{fig_sim1}
\end{figure}

In particular, we present four simulation scenarios consisting of points which lie on: (a) two straight lines; (b) a straight line and a 2-D plane; (c) two 2-D planes; (d) a straight line, a 2-D plane and a 3-D sphere. A typical realisation of each scenarios is illustrated in Figure \ref{fig_sim1}, and for each case we show the original data points in $P$ dimensions (left), the clustering assignments using $K$-means clustering in the original three-dimensional space (centre), and the clustering assignment using PSC (right). It can be noted that the subspaces always intersect so points belonging to different clusters will lie close to each other at these intersections. The $K$-means algorithm, which uses the Euclidean distance between points and has been applied directly to the three-dimensional observations, consistently fails to recover the true clusters, as expected in these cases. On the other hand, PSC correctly recovers both the true clusters and the intrinsic dimensionality of the subspaces. Despite the simplicity of these initial illustrations, they highlight the benefits of subspace clustering. 

A Monte Carlo simulation study was carried out to compare the performance of five competing clustering methods: the proposed PSC algorithm, the $K$-means algorithm (as a benchmark), GPCA \cite{YiMa2006}, $K$-subspaces \cite{Wang2009}, SCC \cite{Chen2008} and Mixture of Common t-Factor Analysers (MCtFA) \cite{Baek2011}. In our studies, in order to make the comparisons fair, we provide each of these methods with the true number of clusters and the true dimensionality of each subspace. In addition to the four simulation settings considered above, we also considered an additional scenario, (e), consisting of four distinct subspaces: a 5-D hyperplane, 4-D hypersphere and two lines embedded in $P=200$ dimensions. Since the dimensionality of this scenario is large, in order to use the GPCA algorithm we must first perform PCA to reduce the dimensionality to $P=7$. 

Results for this comparison are given in Table \ref{tab_sim_nonsparse} which reports on the mean clustering accuracy over $100$ Monte Carlo simulations. 
It can be seen that PSC achieves a consistently high clustering accuracy in all scenarios with a small standard error (reported in parenthesis). SCC also performs extremely well for scenarios (a)-(c) but does more poorly for (d) and (e). The adjusted rand index (ARI) for the same experiments is reported in Table \ref{tab_ari_nonsparse}. In this case, the relative difference in performance between PSC and two other subspace clustering algorithms, GPCA and $K$-subspaces, increases significantly; PCS compares favourably in all scenarios. As before, SCC also shows performance comparable to PSC for (a)-(c), but does less well in (d)-(e). As expected, $K$-means always performs poorly. 

\begin{table}[!t]
	\renewcommand{\arraystretch}{1.3}
	\centering
	\tabcolsep 5.8pt
    \small
	\begin{tabular}{ l l l l l l l}
	\hline
	& \multicolumn{5}{c}{Scenarios} \\	
	Algorithm &  & a & b & c & d & e\\
	\hline 
	PSC  	&   & 0.980 (0.095)  & 0.999 (0.003)  & 1.00 (0.000)  & 0.942 (0.042) & 0.943 (0.089) \\
	GPCA  	&   & 0.846 (0.226)  & 0.786 (0.199)  & 0.921 (0.176)  & 0.791 (0.088) & 0.837 (0.068) \\
	$K$-means & & 0.503 (0.010) & 0.525 (0.028)  &  0.595 (0.097) & 0.635 (0.079) & 0.702 (0.034) \\
	$K$-subspaces & & 0.702 (0.132) & 0.804 (0.151)  &  0.829 (0.149) & 0.766 (0.107) & 0.877 (0.076) \\
	SCC		&  &  0.999 (0.004) & 1.00 (0.000) & 1.00 (0.000) & 0.754 (0.079)& 0.893 (0.073) \\
	MCtFA & & 0.812 (0.158) & 0.984 (0.071)  &  0.883 (0.162) & 0.942 (0.118) & 0.974 (0.049) \\

	 \hline

	\end{tabular}
	\caption{Mean clustering accuracy for six competing clustering algorithms obtained by using five simulated (non-sparse) data sets over $100$ Monte Carlo simulations. Standard errors are reported in parenthesis.}
	\label{tab_sim_nonsparse}
	\end{table}

	\begin{table}[!t]
	\renewcommand{\arraystretch}{1.3}
	\centering
	\tabcolsep 5.8pt
    \small
	\begin{tabular}{ l l l l l l l}
	\hline
	& \multicolumn{5}{c}{Scenarios} \\		
	Algorithm &  & a & b & c & d & e\\
	\hline 
	PSC  	&   & 0.960 (0.189)  & 0.999 (0.006)  & 1.00 (0.000)  & 0.877 (0.181) & 0.943 (0.226) \\
	GPCA  	&   & 0.693 (0.451)  & 0.572 (0.397)  & 0.843 (0.351)  & 0.545 (0.179) & 0.576 (0.171) \\
	$K$-means & & 0.007 (0.021) & 0.055 (0.055)  &  0.190 (0.194) & 0.198 (0.176) & 0.213 (0.093) \\
	$K$-subspaces & & 0.401 (0.263) & 0.609 (0.302)  &  0.659 (0.297) & 0.484 (0.234) & 0.687 (0.186) \\
	SCC		&  &  0.998 (0.008) & 1.00 (0.000) & 1.00 (0.000) & 0.454 (0.175)& 0.719 (0.188) \\
	MCtFA & & 0.625 (0.315) & 0.966 (0.142)  &  0.767 (0.323) & 0.879 (0.246) & 0.936 (0.119) \\

	 \hline

	\end{tabular}
	\caption{Mean ARI for 6 competing clustering algorithms obtained by using five simulated (non-sparse) data sets over $100$ Monte Carlo simulations. Standard errors are reported in parenthesis.}
	\label{tab_ari_nonsparse}
	\end{table}

Finally, we explored the performance of these algorithms on a more challenging set of artificial data. Firstly, compared to the previous examples, we increased the dimensionality of all datasets to $P=200$. Secondly, in order to construct the low-dimensional subspaces, we generated sparse loading vectors where only $10$ randomly chosen variables, out of $200$, had non-zero coefficients, so that the remaining $190$ variables do not contribute to clustering. Gaussian noise with zero-mean and variance $0.5$ was added. Each of the $R_k$ loading vectors in each cluster had the same number of non-zero elements, but the sparsity pattern in each vector was different. This data generating mechanism allows us to test the ability to estimate subspace parameters and simultaneously recover cluster assignments when there is a large number of noisy, irrelevant variables.
Both the true level of sparsity and the true number of clusters are assumed known. For this test case, we compared the performance of PSC using three different levels of sparsity: $10$ (correct sparsity), $100$ and $200$ (no sparsity). We compare to all of the previous methods but replace $K$-means with Sparse $K$-means (SKM), which is also able to perform sparse clustering \cite{Witten2009}, using the correct sparsity level. 

	\begin{table}[!t]
	\renewcommand{\arraystretch}{1.3}
	\centering
	\tabcolsep 5.8pt
    \small
	\begin{tabular}{l l l l l l}
	\hline
	& \multicolumn{4}{c}{Scenarios} \\	
	Algorithm  & a & b & c & d & e\\
	\hline 
	PSC - 10 vars & 0.776 (0.067)  & 0.887 (0.058)  & 0.941(0.034)  & 0.781 (0.045) & 0.812 (0.033) \\
	PSC - 100 vars & 0.731 (0.111)  & 0.740 (0.121)  & 0.549(0.061)  & 0.781 (0.045) & 0.812 (0.031) \\
	PSC - P vars  & 0.643 (0.149)  & 0.708 (0.150)  & 0.533(0.068)  & 0.680 (0.099) & 0.712 (0.041) \\
	GPCA  & 0.500 (0.021)  & 0.500 (0.014)  & 0.524 (0.009)  & 0.634 (0.034) & 0.711 (0.016) \\
	SKM - 10 vars & 0.575 (0.081) & 0.667 (0.137)  &  0.787 (0.147) & 0.590 (0.027) & 0.637 (0.027) \\
	$K$-subspaces & 0.499 (0.005) & 0.499 (0.004)  &  0.498 (0.005) & 0.560 (0.006) & 0.632 (0.005) \\
	SCC	&  0.607 (0.078) & 0.729 (0.091) & 0.832 (0.067) & 0.612 (0.033)& 0.733 (0.034) \\
	MCtFA & 0.596 (0.115) & 0.650 (0.156)  &  0.673 (0.161) & 0.574 (0.091) & 0.663 (0.104) \\

	 \hline

	\end{tabular}
	\caption{Mean clustering accuracy for six competing clustering algorithms obtained by using five simulated (sparse) data sets over $100$ Monte Carlo simulations. Standard errors are reported in parenthesis.}
	\label{tab_sim_sparse}
	\end{table}

	\begin{table}[!t]
	\renewcommand{\arraystretch}{1.3}
	\centering
	\tabcolsep 5.8pt
    \small
	\begin{tabular}{l l l l l l}
	\hline
	& \multicolumn{4}{c}{Scenarios} \\	
	Algorithm & a & b & c & d & e\\
	\hline 
	PSC - 10 vars & 0.551 (0.135)  & 0.774 (0.116)  & 0.883 (0.069)  & 0.509 (0.098) & 0.509 (0.085) \\
	PSC - 100 vars & 0.394 (0.221)  & 0.416 (0.321)  & 0.099 (0.122)  & 0.389 (0.108) & 0.277 (0.086) \\
	PSC - P vars & 0.287 (0.277)  & 0.415 (0.318)  & 0.034 (0.099)  & 0.339 (0.174) & 0.196 (0.073) \\
	GPCA & 0.096 (0.033)  & 0.099 (0.034)  & 0.132 (0.0033)  & 0.174 (0.077) & 0.232 (0.043) \\
	SKM - 10 vars & 0.151 (0.163) & 0.335 (0.275)  &  0.575 (0.292) & 0.080 (0.057) & 0.076 (0.052) \\
	$K$-subspaces & 0.002 (0.009) & 0.003 (0.009)  &  0.004 (0.009) & 0.006 (0.014) & 0.015 (0.013) \\
	SCC	&  0.213 (0.156) & 0.457 (0.182) & 0.663 (0.134) & 0.134 (0.071)& 0.290 (0.091) \\
	MCtFA & 0.194 (0.229) & 0.303 (0.311)  &  0.347 (0.320) & 0.159 (0.135) & 0.284 (0.159) \\

	 \hline

	\end{tabular}
	\caption{Mean ARI for 6 competing clustering algorithms obtained by using five simulated (sparse) data sets over $100$ Monte Carlo simulations. Standard errors are reported in parenthesis.}
	\label{tab_ari_sparse}
	\end{table}

Table \ref{tab_sim_sparse} shows the mean clustering accuracy over $100$ Monte Carlo simulations. When the correct number of informative variables is used, PSC obtains the highest clustering accuracy out of all methods in all scenarios. This is not surprising though, given that the other algorithms to do have any built-in variable selection mechanisms. Amongst the other subspace clustering algorithms, SCC achieves the best results in all scenarios. The unpenalised version of PCS and SCC are directly comparable in this context, and they both achieve similar performances overall, which SCC performing very well for scenario (c). Clearly, when the incorrect number of variables are selected by PSC, its performance decreases. Importantly, PSC performs better than Sparse $K$-means which underlies the importance of performing variable selection in each subspace separately since different variables may be important in each clusters. As we also expect, the performance of PSC degrades as the dimensionality of the subspaces increases. This is due to the constraint that basis vectors of each subspace must be mutually orthonormal. Therefore, if the incorrect sparsity pattern is estimated in the first loading, all subsequent loadings are also likely to be estimated incorrectly. However, since PSC still performs better than all other algorithms in settings (d) and (e) when some sparsity is imposed, this further highlights the benefit of estimating the underlying subspaces using only the truly important variables. 

\begin{table}[!t]
\renewcommand{\arraystretch}{1.3}
\centering
\tabcolsep 5.8pt
 \small
\begin{tabular}{ l l l l l l l}
	\hline
& & \multicolumn{5}{c}{Scenarios} \\
 Algorithm &  & a & b & c & d & e\\
\hline 
SOD - only $K$						&   & 0.86  & 0.66 &  0.90 & 0.45 & 0.35 \\
PRESS - only $K$				&   & 0.89  & 1.0  & 0.96  & 0.62  & 0.70  \\
PRESS - both $K$ and $R_k$ 		&   & 0.73	  & 0.84 & 0.91 & 0.60 & 0.51\\					
 \hline

\end{tabular}
\caption{Performance of PSC used jointly with SOD and the PRESS criterion for learning the number of clusters, $K$. Reported here is the number of times the correct $K$ was selected across $100$ Monte Carlo simulations. The description of the five data sets is in the text.}
\label{tab_sim_ms1}
\end{table}

The mean adjusted rand index reported in Table \ref{tab_ari_sparse} further highlights that this particular setting is particularly challenging for all the algorithms. Here the degradation in performance of PSC when the wrong number of variables are selected appears more clearly. Without sparsity, PSC performs better than other methods in scenarios (a) and (d) and competitively with SCC in (b) and (e). Scenario (c) is particularly challenging and in this case both PSC without any sparsity and K-subspaces perform quite poorly. Here the discriminative subspaces consist of two planes which are corrupted by noise and there is a high degree of overlap between them; moreover, they can only be identified by those 10 relevant variables. However, imposing sparsity in the solution helps PSC identify the correct variables, which in turn produces the best ARI. This scenario highlights one of the main limitations of the unpenalised PSC algorithm which have been overcome by its penalised version. 

Finally, we tested the ability of the PSC algorithm to perform model selection as described in Section \ref{sec_modsel} using the same sparse simulated data as described previously. Table \ref{tab_sim_ms1} shows the number of times that the correct $K$ was selected, in each one of the five scenarios, over $100$ Monte Carlo simulations. Both the PRESS and SOD methods were used to learn the number of clusters. As can be seen in the Table, the PRESS achieves a good performance in the selection of $K$ in the first three scenarios, and its performance decreases when the data contain more clusters. SOD performs similarly in scenarios (a) and (c); however, for all other scenarios, it performs relatively poorly. The bottom row in Table \ref{tab_sim_ms1} shows the performance obtained in learning $K$ when the dimensionality $R_k$ is also learned using the PRESS. Although earning both the dimension of all subspaces and the number of clusters is a more difficult problem, the PRESS still provides satisfactory results. Remarkably, in this more difficult setting, the PRESS is still competitive or even superior than SOD in many scenarios. Some degradation in performance can be explained by the presence of noise causing the algorithm to mistakenly identify 1-dimensional subspaces as 2-dimensional. This is particularly important in setting (a) where the SOD method performs well.

\section{Applications to genomic data sets} \label{real_data}

In order to test the proposed method on real data sets, we applied it to clustering biological samples in six publicly available gene expression datasets. DNA microarrays measure levels of thousands of mRNAs. PCA is routinely used for the analysis and visualisation of these biological measurements because the expression levels of tens of hundreds of genes in samples drawn from the same underlying population are often observed to be highly correlated \cite{Ringner2008}. The individual datasets used for our experiments have been obtained in a preprocessed form from the MIT Broad institute\footnote{{\tt http://www.broadinstitute.org/cgi-bin/cancer/datasets.cgi}}. In each of the datasets, the different classes correspond to different tumour or tissue types relating to different cancers. Each data set is characterised by the number of clusters, ranging from $3$ to $13$, the sample size, $N$, and the number of genes, $P$. A summary of these features is reported in Table \ref{tab_gex}. 

\begin{table}
\begin{center}
	\tabcolsep 5.8pt
    \small
	\begin{tabular}{llrrr}
\hline
&Dataset & $K$ & $N$ & $P$ \\ \hline
1&Leukemia \cite{Golub1999} & 3 & 38 & 999 \\
2&CNS tumors \cite{Pomeroy2001} & 5 & 48 & 1000 \\
3&Normal tissues \cite{Ramaswamy2001} & 13 & 99 & 1277 \\
4&St. Jude leukemia \cite{Yeoh2002} & 6 & 248 & 985 \\
5&Lung cancer \cite{Bhattacharjee2001} & 4 & 197 & 1000 \\
6&Novartis multi-tissue \cite{Su2002} & 4 &103 & 1000 \\
\hline
\end{tabular}
\caption{Six gene expression data sets. \label{tab_gex}}
\end{center}
\end{table}

We compared the clustering performance obtained by PSC against the other state-of-art subspace clustering algorithms that were tested on simulated data sets. Of all the algorithms considered, both Mixture of Common t-Factor Analysers (MCtFA) and Sparse $K$-means (SKM) have been previously applied to clustering gene expression data, and SKM in particular is able to perform gene selection. The PSC algorithm was run using up to $3$ latent factors, and $5$ different levels of sparsity, where the number of selected genes is between $10$ and $P$. Clearly, when $P$ genes are selected, no sparse solutions are imposed. The SKM algorithm is run using the same levels of sparsity and, even in this case, selecting $P$ variables is equivalent to standard $K$-means. For $K$-subspaces and SKM we use $50$ random restarts and take the cluster configuration which minimises the within cluster sum of squares. SCC is run $50$ times, each time using up to $R=3$ dimensions and take the mean result. MCtFA is run with $50$ restarts, the results which minimise the negative log likelihood are taken as final. 

In Tables \ref{tab_gex_results} and \ref{tab_ari_results} we report on the clustering accuracy and Adjusted Rand Index (ARI), respectively. Highlighted in each column is the best performance attained for each method. The PSC algorithm without sparsity and with two latent factors achieves a good clustering accuracy on all but one of the datasets indicating that samples belonging to different clusters can be well approximated by a two-dimensional linear subspace; these results also confirm that PSC is able to estimate this subspace accurately in high-dimensions. It can be seen that in all cases, maximum clustering accuracy can be achieved when some degree of sparsity is imposed. When $R=1$, the best solution is achieved when $50$ or fewer variables have been selected. For larger values of $R$, the best solution occurs when $100$ or fewer variables have been selected. For all values of $R$, the non-sparse PSC algorithm never outperforms the best sparse PSC, which indicates that a certain number of genes in these data sets have a small contribution towards the determination of the various clusters. 

It is important to observe that standard (non-penalised) $K$-means performs almost equivalently to PSC on datasets $1$ and $3$ indicating that these clusters are well separated geometrically. However, as the level of sparsity is increased, the accuracy of $K$-means typically decreases. This is because the sparsity is estimated using all samples and selected genes are necessarily present in all clusters. PSC achieves better performance because it selects the important genes within each cluster individually. Compared to $K$-means, SCC achieves a greater clustering accuracy on all but datasets $1$ and $3$. Increasing the dimensionality of the subspaces does not greatly affect the clustering accuracy. The large difference in clustering performance on dataset $1$ between SCC and PSC seems to suggest that PSC may perform better in higher dimensions.

When one a one-dimensional subspace is extracted, $K$-subspaces achieves a good clustering accuracy and outperforms standard PSC on datasets $1$,$3$ and $4$. However, on these datasets, $K$-subspaces is also extremely sensitive to the number of dimensions. In all cases, the clustering performance decreases monotonically when the number of dimensions is increased, which makes the problem of selecting the best model particularly difficult. GPCA displays similar performance to $K$-subspaces, whereas MCtFA typically performs worse than PSC for all values of $R$. However for $R>1$, MCtFA performs equivalently to PSC on datasets $4$, $5$ and $6$. 

The effect of changing the number of latent factors in PSC, SCC, MCtFA and GPCA can also be compared. For SCC, the effect on the clustering accuracy of changing $R$ is typically not large. However, for PSC and MCtFA, the effect of changing the subspace dimensionality generally varies between datasets and is non-monotonic. This non-monotonic behaviour was also observed by \cite{Baek2011} for values of up to $R=10$. As noted before, this makes model selection particularly challenging. It could be argued that, for some datasets, increasing the number of latent factors beyond three might be helpful for some algorithms; although this may be the case, increasing the number of latent factors in each clyster requires the estimation of a much larger set of parameters, and this seems unnecessary in light of the good performance of PSC, GPCA and $K$-subspaces with only a few latent factors. As the number latent factors increases, we also observe that the performance of GPCA decreases. The corresponding results using ARI as a performance measure highlights a larger disparity between competing methods and PSC, in support for the latter. In particular, the results suggest that these other methods may miss smaller clusters. This is particularly evident in dataset $3$ which is especially challenging due to the presence of $13$ clusters of varying sizes.

We also attempted to learn the number of clusters in the six gene expression datasets using the PRESS based method described in Section \ref{sec_modsel} by setting $R=1$ and selecting 10 variables. The results are reported in Table \ref{tab_ms_gex}. It can be seen that the PRESS is able to correctly identify the true number of clusters in three of the datasets. In datasets 4 and 5 the number of clusters is underestimated by one. However, the PRESS is unable to correctly identify the true number of clusters in dataset 3. This is perhaps not surprising because this data set contains clusters of normal tissues, which are quite similar to each other, whereas the other datasets contain sub-types of cancerous tissues which show a much higher degree of separation.


	\begin{table}
	\begin{center}
		\tabcolsep 5.8pt
	    \small	
	\begin{tabular}{lrrrrrrr}
	\hline
	& & & \multicolumn{5}{c}{Data Set} \\	
	Algorithm &                    		& 1          	& 2 		& 3		& 4 		& 5 		& 6  		\\ \hline
	\multirow{5}{*}{PSC ($1$)}&10& 0.959	& {\bf 0.932}	& 0.910	& {\bf 0.969}    &{\bf 0.922}	&{\bf 0.954}	\\
					& 50		& {\bf 0.963}	& 0.929	& {\bf 0.938}	& 0.950	& 0.888	& 0.944	\\
					& 100	& 0.936	& 0.887	& 0.926	& 0.946	& 0.855	& 0.903	\\
					& 500	& 0.963	& 0.866	& 0.908	& 0.873	& 0.707	& 0.849	\\
					& P 		& 0.862	& 0.830	& 0.915	& 0.875	& 0.695	& 0.833	\\ 
	\\
	\multirow{5}{*}{PSC ($2$)} &10& 0.911	& 0.905	& 0.923	& 0.963	& 0.815  & {\bf 0.989}	\\
					& 50		& {\bf 1.000}	& {\bf 0.914}	& {\bf 0.929}	& 0.961	& 0.839	& 0.980	\\
					& 100	& 1.000	& 0.892	& 0.923	&{\bf 0.968}	&{\bf 0.866}	& 0.980	\\
					& 500	& 1.000	& 0.858	& 0.906	& 0.949	& 0.715	& 0.980	\\
					& P 		& 1.000	& 0.893	& 0.918	& 0.961	& 0.621	& 0.980	\\ 
	\\
	\multirow{5}{*}{PSC ($3$)} &10& 0.959	& {\bf 0.894}	& 0.871	& 0.962	& 0.858  & {\bf 0.990}	\\
					& 50		& {\bf 1.000}	& 0.861	& 0.911	& 0.969	& 0.845	& 0.990	\\
					& 100	& 0.959	& 0.877	& {\bf 0.927}	&{\bf 0.980}	&{\bf 0.865}	& 0.990	\\
					& 500	& 0.963	& 0.850	& 0.879	& 0.972	& 0.816	& 0.990	\\
					& P 		& 0.923	& 0.854	& 0.876	& 0.953	& 0.740	& 0.990	\\ 
	\\
	\multirow{5}{*}{Sparse $K$-means}	&10	& 0.812	& 0.695	& 0.871	& 0.774	& 0.547	& 0.755	\\
	 							&50	& 0.852	& 0.720	& 0.898	&{\bf 0.777}	& 0.580	& {\bf 0.765}	\\
								&100& 0.885	& {\bf 0.734}	& 0.913	& 0.776	& 0.580	& 0.765	\\
								&500& {\bf 0.959}	& 0.734	& {\bf  0.916}	& 0.776	&{\bf 0.582}	& 0.765	\\
								&P	& 0.959	& 0.734	& 0.913	& 0.776	& 0.582	& 0.765	\\
	\\
	$K$-subspaces ($1$)	&	&{\bf 0.918} & {\bf 0.784} & {\bf 0.932} & {\bf 0.935} & {\bf 0.660} & {\bf 0.819} \\
	$K$-subspaces ($2$)	&	&0.684 &	0.761 & 0.921 & 0.814 & 0.593 & 0.747 \\
	$K$-subspaces ($3$)	&	&0.558 &	0.718 & 0.877 & 0.771 & 0.563 & 0.715 \\
	\\
	SCC ($1$) 			 &   	& {\bf 0.675}	& 0.770	& {\bf 0.865}	&{\bf 0.869}	& 0.517	& 0.897 	\\
	SCC ($2$)				 &   	& 0.671	& 0.772	& 0.864	& 0.868	& 0.594	& 0.898	\\
	SCC ($3$)			 	&   	& 0.672	& {\bf 0.774}	& 0.865	& 0.869	&{\bf 0.598}	& {\bf 0.901}	\\
	\\
	MCtFA ($1$)			&  	& 0.652	&{\bf  0.717}	& 0.630	& 0.820	& 0.645	& 0.696	\\
	MCtFA ($2$)			&  	& 0.693	& 0.522	& 0.642	&{\bf 0.932}	& 0.732	& 0.943	\\
	MCtFA ($3$)			&  	& {\bf 0.879}	& 0.689	&{\bf 0.725}	& 0.862	&{\bf 0.922}	& {\bf 0.963}	\\
	\\
	GPCA ($1$)			&  	& {\bf 0.852}	&{\bf  0.836}	&{\bf  0.912}	&{\bf  0.905}	&{\bf  0.580}	&{\bf  0.803}	\\
	GPCA ($2$)			&  	&{ 0.822}	&{ 0.799}	& 0.903	&{ 0.759}	& 0.562	&{ 0.708}	\\
	GPCA ($3$)			&  	& 0.590	& 0.706	&{ 0.866}	&{0.755}	&{ 0.558}	& 0.746	\\
	\\

	\hline
	\end{tabular}
	
		\caption{Mean clustering accuracy of competing clustering algorithms on the six gene expression datasets described in table \ref{tab_gex}.\label{tab_gex_results}}
	\end{center}
	\end{table}
	

	\begin{table}
	\begin{center}
		\tabcolsep 5.8pt
	    \small
	\begin{tabular}{lrrrrrrr}
	\hline
	& & & \multicolumn{5}{c}{Data Set} \\	
	Algorithm &                    		& 1          	& 2 		& 3		& 4 		& 5 		& 6  		\\ \hline
	\multirow{5}{*}{PSC ($1$)}&10& 0.911	&{\bf 0.778}	& 0.400	&{\bf 0.909}    &{\bf 0.844}	&{\bf 0.874}	\\
					& 50		&{\bf 0.919}	& 0.772	&{\bf 0.586}	& 0.850	& 0.776	& 0.849	\\
					& 100	& 0.860	& 0.655	& 0.507	& 0.841	& 0.712	& 0.739	\\
					& 500	& 0.919	& 0.578	& 0.407	& 0.619	& 0.424	& 0.612	\\
					& P 		& 0.698	& 0.460	& 0.434	& 0.636	& 0.401	& 0.547	\\ 
	\\
	\multirow{5}{*}{PSC ($2$)} &10& 0.911	& 0.697	& 0.495	& 0.889	& 0.634  &{\bf  0.973}	\\
					& 50		& {\bf 1.000}	&{\bf 0.732}	&{\bf 0.542}	& 0.881	& 0.681	& 0.946	\\
					& 100	& 1.000	& 0.667	& 0.466	&{\bf 0.906}	&{\bf 0.735}	& 0.946	\\
					& 500	& 1.000	& 0.577	& 0.440	& 0.849	& 0.441	& 0.946	\\
					& P 		& 1.000	& 0.673	& 0.493	& 0.885	& 0.639	& 0.946	\\ 
	\\
	\multirow{5}{*}{PSC ($3$)} &10& 0.911	&{\bf 0.675}	& 0.038	& 0.888	& 0.719  &{\bf  0.973}	\\
					& 50		&{\bf 1.000}	& 0.570	& 0.370	& 0.907	& 0.693	& 0.973	\\
					& 100	& 0.959	&{0.613}	&{\bf 0.496}	&{\bf 0.942}	&{\bf 0.731}	& 0.973	\\
					& 500	& 0.963	& 0.533	& 0.293	& 0.917	& 0.637	& 0.973	\\
					& P 		& 0.923	& 0.548	& 0.131	& 0.865	& 0.488	& 0.973	\\ 
	\\
	\multirow{5}{*}{Sparse $K$-means}	&10	& 0.605	& 0.128	& 0.236	& 0.342	& 0.115	&{\bf 0.426}	\\
	 							&50	& 0.685	&{0.203}	& 0.379	& 0.345	& 0.176	& 0.421	\\
								&100& 0.749	& {\bf 0.212}	& 0.435	&{\bf 0.355}	& 0.176	& 0.421	\\
								&500&{\bf 0.911}	& 0.212	&{\bf 0.457}	& 0.348	&{\bf 0.181}	& 0.421	\\
								&P	& 0.911	& 0.212	& 0.455	& 0.348	& 0.181	& 0.421	\\
	\\
	$K$-subspaces ($1$)	&	&{\bf 0.824} & {\bf 0.424} & {\bf 0.682} & {\bf 0.803} & {\bf 0.333} & {\bf 0.544} \\
	$K$-subspaces ($2$)	&	&0.320 &	0.269 & 0.490 & 0.431 & 0.205 & 0.356 \\
	$K$-subspaces ($3$)	&	&0.030 &	0.141 & 0.289 & 0.295 & 0.140 & 0.248 \\
	\\
	SCC ($1$) 			 &   	& 0.291	& 0.302	& 0.173	& 0.579	&{\bf 0.215}	& 0.717 	\\
	SCC ($2$)				 &   	& 0.289	& 0.299	& 0.183	&{\bf 0.581}	& 0.210	&{\bf 0.730}	\\
	SCC ($3$)			 	&   	&{\bf 0.310}	&{\bf 0.316}	&{\bf 0.193}	& 0.571	& 0.203	& 0.705	\\
	\\
	MCtFA ($1$)			&  	& 0.367	& 0.226	& 0.034	& 0.708	& 0.149	& 0.429	\\
	MCtFA ($2$)			&  	&{\bf 0.830}	&{\bf 0.317}	& 0.055	&{\bf 0.809}	& 0.598	&{\bf 0.921}	\\
	MCtFA ($3$)			&  	& 0.748	& 0.213	&{\bf 0.085}	&{0.759}	&{\bf 0.844}	& 0.899	\\

	\\
	GPCA ($1$)			&  	&{\bf 0.692}	&{\bf  0.540}	&{\bf  0.451}	& {\bf 0.725}	& {\bf 0.181}	&{\bf  0.514}	\\
	GPCA ($2$)			&  	&{0.607}	&{ 0.373}	& 0.339	&{ 0.424}	& 0.143	&{ 0.261}	\\
	GPCA ($3$)			&  	& 0.112	& 0.087	&{ 0.09}	&{0.318}	&{ 0.133}	& 0.372	\\	
	\\
	\hline
	\end{tabular}

		\caption{Mean ARI of competing clustering algorithms on the six gene expression datasets described in table \ref{tab_gex}.\label{tab_ari_results}}
	\end{center}
	\end{table}

\begin{table}
\begin{center}
	\tabcolsep 5pt
    \small
\begin{tabular}{lrr}
\hline
Dataset & Real $K$ & Estimated $K$ \\ \hline
1& 3 & 3 \\
2 & 5 & 5 \\
3& 13 & 6 \\
4& 6 & 5 \\
5& 4 & 5 \\
6& 4 & 4 \\
\hline
\end{tabular}
\caption{Real vs learned number of data clusters obtained by PCS with the PRESS using only the first principal component on the six gene expression data sets. \label{tab_ms_gex}}
\end{center}
\end{table}

\section{Conclusion} \label{sec_conc}

In this work we have introduced an efficient approach to subspace clustering for high-dimensional data. Our algorithm relies on a new measure of influence for PCA, derived from an approximated PCA PRESS statistic, which can also be used in other applications unrelated to clustering to detect influential observations. Compared to our initial work in \cite{McWilliams2011}, here we have presented the relevant methodology in much greater detail, along with a number of extensions and additional empirical evaluations, including a proof of convergence of the PSC algorithm. 

A penalised version of the algorithm has also been introduced that can perform simultaneous subspace clustering and variable selection. For high-dimensional data, penalising the PCA solution in this way aids in the interpretability of the resulting partitions by identifying which variables contribute to each latent factor, and therefore which variables are important for explaining the directions of maximal variability in the data. Although the problem of variable selection in clustering has been discussed before (see, for instance, \cite{Witten2009}), we are not aware of other subspace clustering algorithms which estimate sparse subspace parameters. Furthermore, more structured penalties could be used instead of the simple $l_1$ penalty \cite{Friedman2007}. For instance, non-negative cluster-wise parameters may be more appropriate in some situations \cite{Witten2010}.

Extensive simulation experiments have been presented here that compare a number of subspace clustering algorithms. For these experiments, we have selected challenging scenarios in which the data within each cluster have low-dimensional representations that need to be identified, including the case where these representations are sparse. Our results indicate the SPC is particularly competitive in a wide range of situations. Apart from comparisons on artificially constructed data, we have also tested whether a subspace approach is beneficial in clustering biological samples for which a thousand gene expression levels have been measured. Our empirical evaluation using six different publicly available data sets suggest that, although the clusters in some data sets might be discovered using more traditional algorithms that exploits geometric structures, subspace clustering is very useful in many other datasets due to the fact that gene expressions levels within each cluster can be approximated well by a few principal components. In our experiments, PSC always appears to be very competitive and, in several situations, has also been shown to out-performs other competing methods. Apart from gene expression data, the PSC algorithm had been previously shown to perform particularly well in clustering digital images of human faces collected under different lighting conditions for which a cluster-wise PCA reconstruction is also appropriate \cite{McWilliams2011}.


The PSC algorithm maintains some similarity to the $K$-subspaces algorithm. As with $K$-subspaces, we iteratively fit cluster-wise PCA models and reassign points to clusters until a certain optimality condition is met. However, rather than trying to minimise the residuals under the individual PCA models, we introduce an objective function that exploits the predictive nature of PCA in a way that makes it particularly robust against overfitting.  Along with the PRESS statistic, the PSC is able to learn both the number of clusters and the dimensionality of each subspace, although this is a particularly difficult problem and more investigation is required. The difficult issue of selecting the correct level of sparsity within each subspace will also be explored further in further work. 


\appendix
\section{Derivation of predictive influence} \label{sec_appA}

Using the chain rule, the gradient of the PRESS for a single latent factor is
$$
\frac{\partial J^{(1)}}{\partial \bm{x}_i} =  \frac{1}{2} \dx \lv \bm{e}^{(1)}_{-i}\rv^2  = \frac{1}{2}\bm{e}^{(1)}_{-i} \dx \bm{e}^{(1)}_{-i}.
$$
For notational convenience we drop the superscript in the following. 
Using the quotient rule, the partial derivative of the $i^{th}$ leave-one-out error has the following form
\begin{align*}
 \frac{\partial  }{ \partial \bm{x}_i} \bm{e}_{-i}  =  \frac{\frac{\partial   }{ \partial \bm{x}_i} \bm{e}_i (1-h_i) +  \bm{e}_i\frac{\partial  h_i }{ \partial \bm{x}_i}}{(1-h_i)^2}
\end{align*}
which depends on the partial derivatives of the $i^{th}$ reconstruction error and the $h_i$ quantities with respect to the observation $\bm{x}_i$. The computation of these two partial derivatives are straightforward and are, respectively
$$
\frac{\partial  }{ \partial \bm{x}_i} \bm{e}_i  
= \frac{\partial}{\partial \bm{x}_i}  \bm{x}_i \left(\bm{I}_P - \bm{v}{\bm{v}}\tr\right) 
= \left(\bm{I}_P - \bm{v}{\bm{v}}\tr\right) ,
$$
and
$$
\frac{\partial  }{ \partial \bm{x}_i}  h_i = \frac{\partial}{\partial \bm{x}_i} 
\bm{x}_i\bm{v} D \bm{v}\tr \bm{x}_i\tr = 2\bm{v} D d_i .
$$
The derivative of the PRESS, $J$ with respect to $\bm{x}_i$ is then
\begin{equation}
\frac{1}{2}\frac{\partial  }{ \partial \bm{x}_i}\lv \bm{e}_{-i}\rv^2  = \bm{e}_{-i} \frac{\partial   }{ \partial \bm{x}_i} \bm{e}_{-i}= \bm{e}_{-i}  \frac{ \left(\bm{I}_P - \bm{v}{\bm{v}}\tr\right) (1-h_i) +  2\bm{e}_i \bm{v} D d_i }{(1-h_i)^2}.
\label{eq_dPRESS}
\end{equation}
However, examining the second term in the sum, $\bm{e}_i \bm{v} D d_i $, we notice
$$
\bm{e}_i\bm{v}Dd_i = (\bm{x}_i-\bm{x}_i\bm{vv}\tr)\bm{v}Dd_i 
									 = \bm{x}_i\bm{v}Dd_i - \bm{x}_i\bm{vv}\tr \bm{v}Dd_i
									 = 0 . 
$$
Substituting this result back in Eq. \eqref{eq_dPRESS}, the gradient of the PRESS for a single PCA component with respect to $\bm{x}_i$ is given by
$$
\frac{1}{2}\frac{\partial  }{ \partial \bm{x}_i} \lv \bm{e}_{-i}\rv^2 = \bm{e}_{-i} 
  \frac{ \left(\bm{I}_P - \bm{v}{\bm{v}}\tr\right) (1-h_i)}{(1-h_i)^2} 
  = \bm{e}_{-i} 
  \frac{ \left(\bm{I}_P - \bm{v}{\bm{v}}\tr\right) }{(1-h_i)} .
$$

In the general case for $R>1$, the final expression for the predictive influence $\bm{\pi}(\bm{x}_i)\R^{P\times 1}$ of a point $\bm{x}_i$ under a PCA model then has the following form:
$$
\bm{\pi}(\bm{x}_i;\bm{V})  = \bm{e}^{(R)}_{-i}
  \left( \sum_{r=1}^{R} \frac{ \left(\bm{I}_p - \bm{v}^{(r)}{\bm{v}^{(r)}}\tr\right) }{\left(1-h^{(r)}_i\right)} - (R-1) \right) .
$$

	
	\section{Proof of Lemma \ref{lem_pca_maxprob}} \label{proof_pca_maxprob}

	From Appendix \ref{sec_appA}, for $R=1$, the predictive influence of a point $\bm{\pi}({\bm{x}_i};\bm{v})$ is 
	\begin{align}
	\bm{\pi}(\bm{x}_i;\bm{v}) 
	&=\frac{\bm{e}_{i}}{(1-h_i)^2}
	\label{eq_piR1}
	\end{align}
	This is simply the $i^{th}$ leave-one-out error scaled by $1-h_i$. If we define a diagonal matrix $\bm{\Xi}\R^{N\times N}$ with diagonal entries ${\Xi}_{i} = (1-h_i)^2$, we can define a matrix $\bm{\Pi}\R^{N\times P}$ whose rows are the predictive influences, $\bm{\Pi}=[\bm{\pi}(\bm{x}_1;\bm{v}) \tr ,\ldots, \bm{\pi}(\bm{x}_N;\bm{v}) \tr]\tr$. This matrix has the form
	\[
	\bm{\Pi} = \bm{\Xi}^{-1}\left(\bm{X} - \bm{X}\bm{vv}\tr\right) .
	\]
	Now, solving \eqref{eq_rpca_opt} is equivalent to minimising the squared Frobenius norm, 
	\begin{align}
	\label{eq_min_predinf_trace}
	&\min_{\bm{v}} \text{Tr} \left(\left(\bm{X} - \bm{X}\bm{vv}\tr\right)\tr \bm{\Xi}^{-2} \left(\bm{X} - \bm{X}\bm{vv}\tr\right)\right) \\
	&\text{subject to } ~ \lv \bm{v} \rv =1 . \nonumber
	\end{align}
	Expanding the terms within the trace we obtain
	\begin{align*}
	& \text{Tr} \left(\left(\bm{X} - \bm{X}\bm{vv}\tr\right)\tr \bm{\Xi}^{-2} \left(\bm{X} - \bm{X}\bm{vv}\tr\right)\right) \nonumber\\
	& = \text{Tr} \left( \bm{X}\tr \bm{\Xi}^{-2} \bm{X} \right) - 2\text{Tr}\left(\bm{vv}\tr \bm{X} \tr \bm{\Xi}^{-2} \bm{X} \right) +
	\text{Tr}\left(\bm{vv}\tr \bm{X} \tr \bm{\Xi}^{-2} \bm{X}\bm{vv}\tr \right).
	\end{align*}
	By the properties of the trace, the following equalities hold
	\begin{align*}
	\text{Tr}\left(\bm{vv}\tr \bm{X} \tr \bm{\Xi}^{-2} \bm{X} \right) = \bm{v}\tr \bm{X}\tr \bm{\Xi}^{-2} \bm{X} \bm{v},
	\end{align*}
	and
	\begin{align*}
	\text{Tr} \left( \bm{vv}\tr \bm{X} \tr \bm{\Xi}^{-2} \bm{X}\bm{vv}\tr\right) & = \text{Tr}\left(\bm{\Xi}^{-1}\bm{X}\bm{vv}\tr \bm{vv}\tr\bm{X}\tr\bm{\Xi}^{-1}\right) \nonumber\\
	& = \bm{v}\tr \bm{X}\tr \bm{\Xi}^{-2} \bm{X} \bm{v},
	\end{align*}
	since $\bm{\Xi}$ is diagonal and $\bm{v}\tr\bm{v}=1$.
	Therefore, \eqref{eq_min_predinf_trace} is equivalent to
	\begin{align}
	\label{eq_weighted_ev2}
	&\min_{\bm{v}} \text{Tr} \bm{X}\tr \bm{\Xi}^{-2} \bm{X} -  \bm{v}\tr\bm{X}\tr\bm{\Xi}^{-2}\bm{Xv} , \\
	&\text{subject to } ~ \lv \bm{v} \rv =1 . \nonumber
	\end{align}
	It can be seen that under this constraint, Eq. \eqref{eq_weighted_ev2} is minimised when $\bm{v}\tr\bm{X}\tr\bm{\Xi}^{-2}\bm{Xv}$ is maximised which, for a fixed $\bm{\Xi}$ is achieved when $\bm{v}$ is the eigenvector corresponding to the largest eigenvalue of  $\bm{X}\tr \bm{\Xi}^{-2} \bm{X}$.

	\section{Proof of Lemma \ref{lem_pca_svd}}
	\label{proof_pca_svd}
	In this section we provide a proof of Lemma \ref{lem_pca_svd} 
	As an additional consequence of this proof, we develop an upper bound for the approximation error which can be shown to depend on the leverage terms. We derive this result for a single cluster, $\mathcal{C}^{(\tau)}$ however it holds for all clusters.

	We represent the assignment of points $i=1,\ldots,N$ to a cluster, $\mathcal{C}^{(\tau)}$ using a binary valued diagonal matrix $\bm{A}$ whose diagonal entries are given by
	\begin{equation}
	A_{i}= \left\{ \begin{array}{ll} 1, & \text{ if } i\in\mathcal{C}^{(\tau)} \\ 
	0,& \text{ otherwise},
	\end{array} \right.
	\end{equation}
	where $\text{Tr}(\bm{A})=N_k$.
	We have shown in Lemma \ref{lem_pca_maxprob} that for a given cluster assignment, the parameters which optimise the objective function can be estimated by computing the SVD of the matrix
	\begin{align}
	\sum_{i\in \mathcal{C}_k^{(\tau)}} \bm{x}_i\tr {\Xi}_{i}^{-2} \bm{x}_i = \bm{X}\tr \bm{\Xi}^{-2}\bm{A} \bm{X} ,
	\label{eq_pls_optmat_pca}
	\end{align}
	within each cluster where the $i^{th}$ diagonal element of ${\bm{\Xi}}$ is $\Xi_{i}=(1-h_i)^2\leq1$, so that $\Xi_{i}^{-2}\geq 1$. We can then represent ${\bm{\Xi}}^{-2} = \bm{I}_N + \bm{\Phi}$ where $\bm{\Phi}\R^{n\times n}$ is a diagonal matrix with entries $\Phi_{i}=\phi_i\geq 0$. 
	Now, we can represent Eq. \eqref{eq_pls_optmat_pca} at the next iteration as
	%
	\begin{align}
	\bm{M} = & \bm{X}\tr\bm{A}(\bm{I}_N + \bm{\Phi})\bm{X} .
	\end{align}

	We can quantify the difference between the optimal parameter, $\bm{v}^{*}$ obtained by solving \eqref{eq_pca_max_pi} using $ \bm{M}$ and the new PCA parameter estimated at iteration $\tau+1$,  $\bm{v}^{(\tau)}$ as, 
	$$
	E(\mathcal{S}^*,\mathcal{S}^{(\tau)})= {\bm{v}^{*}}\tr \bm{M}^{(\tau)} \bm{v}^{*} - {\bm{v}^{(\tau)}}\tr \bm{X}\tr \bm{A} \bm{X}\bm{v}^{(\tau)},
	$$
	where $\bm{v}^{(\tau)}$ is obtained through the SVD of $ \bm{X}\tr\bm{A}\bm{X} $. We can express $E(\mathcal{S}^*,\mathcal{S}^{(\tau)})$ in terms of the spectral norm of $\bm{M}$. Since the spectral norm of a matrix is equivalent to its largest singular value, we have ${\bm{v}^{(\tau)}}\tr \bm{X}\tr \bm{A} \bm{X}\bm{v}^{(\tau)} =\lv  \bm{X}\tr\bm{A}\bm{X} \rv$
	Since $\bm{\Phi}$ is a diagonal matrix, its spectral norm, $\lv \bm{\Phi} \rv = \max(\bm{\Phi})$. Similarly, $\bm{A}$ is a diagonal matrix with binary valued entries, so $\lv \bm{A} \rv = 1$. 
	\begin{align}
	E(\mathcal{S}^*,\mathcal{S}^{(\tau)}) & \leq  \lv \bm{M} - \bm{X}\tr \bm{A}\bm{X} \rv \nonumber\\
	& \leq \lv \bm{X}\tr\bm{A}\bm{\Phi}\bm{X} \rv \nonumber \\
	& \leq \max(\bm{\Phi}) \lv \bm{X}\tr\bm{X} \rv .
	\label{eq_plsE2_pca}
	\end{align}
	Where the triangle and Cauchy-Schwarz inequalities have been used.
	In a similar way, we now quantify the difference between the optimal parameter and the old PCA parameter $\bm{v}^{(\tau-1)}$,
	$$
	E(\mathcal{S}^*,\mathcal{S}^{(\tau-1)}) = {\bm{v}^{*}}\tr \bm{M} \bm{v}^{*} - {\bm{v}^{(\tau-1)}}\tr \bm{X}\tr \bm{A} \bm{X}\bm{v}^{(\tau-1)}.
	$$
	Since $\bm{v}^{(\tau)}$ is the principal eigenvector of $\bm{X}\tr\bm{A}\bm{X}$, by definition, ${\bm{v}^{(\tau)}}\tr\bm{X}\tr\bm{A}\bm{Xv}^{(\tau)}$ is maximised, therefore we can represent the difference between the new parameters and the old parameters as
	$$E(\mathcal{S}^{(\tau)},\mathcal{S}^{(\tau-1)})={\bm{v}^{(\tau)}}\tr\bm{X}\tr\bm{A} \bm{Xv}^{(\tau)} - {\bm{v}^{(\tau-1)}}\tr\bm{X}\tr\bm{A} \bm{Xv}^{(\tau-1)}\geq 0.$$ 
	Using this quantity, we can express $E(\mathcal{S}^*,\mathcal{S}^{(\tau-1)})$ as
	\begin{align}
	E(\mathcal{S}^*,\mathcal{S}^{(\tau-1)})  \leq & \lv \bm{M}  \rv -  {\bm{v}^{(\tau-1)}}\tr \bm{X}\tr \bm{A} \bm{X}\bm{v}^{(\tau-1)} \nonumber\\
	\leq & \lv \bm{X}\tr\bm{\Phi} \bm{A} \bm{X} \rv + \lv \bm{X}\tr \bm{A} \bm{X} \rv -   {\bm{v}^{(\tau-1)}}\tr \bm{X}\tr \bm{A}  \bm{X}\bm{v}^{(\tau-1)} \nonumber\\
	\leq & \max(\bm{\Phi} )\lv \bm{X}\tr\bm{X}  \rv  + E(\mathcal{S}^{(\tau)},\mathcal{S}^{(\tau-1)}),
	\label{eq_plsE1_2_pca}
	\end{align}

	From Eq. \eqref{eq_plsE1_2_pca} and \eqref{eq_plsE2_pca} it is clear that
	\begin{align}
	E(\mathcal{S}^*,\mathcal{S}^{(\tau)}) \leq E(\mathcal{S}^*,\mathcal{S}^{(\tau-1)}) .
	\label{eq_pca_anglediff}
	\end{align}
	This proves Lemma \ref{lem_pca_svd}.

	The inequality in Eq. \eqref{eq_pca_anglediff} implies that estimating the SVD using $\bm{X}\tr\bm{A} \bm{X}$ obtains PCA parameters which are closer to the optimal values than those obtained at the previous iteration. Therefore, estimating a new PCA model after each cluster re-assignment step never increases the objective function. Furthermore, as the recovered clustering becomes more accurate, by definition there are fewer influential observations within each cluster. This implies that $\max(\bm{\Phi} ) \rightarrow 0$, and so $ E(\mathcal{S}^*,\mathcal{S}^{(\tau)}) \rightarrow0$.

	


\bibliographystyle{spmpsci}      
\bibliography{ppsc}

\end{document}